\documentclass[runningheads]{llncs}

\usepackage{eccv}

\usepackage{eccvabbrv}

\usepackage{graphicx}
\usepackage{booktabs}

\usepackage{url}

\usepackage{graphicx}
\usepackage{makecell}
\usepackage{booktabs}
\usepackage{multirow}
\usepackage{multicol}
\usepackage{algorithm}
\usepackage{algorithmicx}
\usepackage{colortbl}
\usepackage{amssymb}
\usepackage{ntheorem}
\usepackage{wrapfig}

\usepackage{amsmath}

\definecolor{mygray}{gray}{.9}
\definecolor{mypink}{rgb}{.99,.91,.95}
\definecolor{mygreen}{rgb}{.52,.73,.30}
\definecolor{myblue}{rgb}{.39,.58,.93}
\definecolor{mycyan}{cmyk}{.3,0,0,0}
\definecolor{mylakeblue}{rgb}{.0,.749,1.}
\definecolor{mypurple}{rgb}{.729,.333,.827}
\definecolor{myclassicblue2}{rgb}{.117,.565,1.}
\definecolor{myclassicblue}{rgb}{1.0,.3176,.3176}
\definecolor{myorange}{rgb}{1.0,.647,0}
\definecolor{kanki}{rgb}{.941,.902,.549}
\definecolor{mybrown}{rgb}{0.7176,0.4313,0}
\definecolor{mybrown2}{rgb}{0.57254,0.3490,0.0078}
\definecolor{myblue3}{rgb}{0.1019,0.1372,0.4941}
\definecolor{manifoldpurple}{rgb}{0.8509803921568627, 0.5372549019607843, 0.8117647058823529}
\definecolor{manifoldblue}{rgb}{0.615686274509804, 0.7647058823529411, 0.9019607843137255}
\definecolor{manifoldgreen}{rgb}{.6627450980392157, 0.8196078431372549, 0.5568627450980392}

\definecolor{myredddd}{rgb}{1.0, 0.3176470588235294, 0.3176470588235294}
\definecolor{mypurpleee}{rgb}{0.3607843137254902,0.47843137254901963,0.9176470588235294}

\newcommand{\bred}[1]{\textcolor{red}{\textbf{#1}}}
\newcommand{\iblue}[1]{\textcolor{blue}{\textit{#1}}}

\usepackage[noend]{algpseudocode} 

\usepackage[pagebackref=true,breaklinks=true,letterpaper=true,colorlinks,bookmarks=false,citecolor=myblue3]{hyperref}

\usepackage{orcidlink}

\begin{document}

\title{Parameterized Quasi-Physical Simulators for Dexterous Manipulations Transfer} 

\titlerunning{Quasi-Physical Simulators for Dexterous Manipulations Transfer}

\author{Xueyi Liu\inst{1,3}
\and
Kangbo Lyu\inst{1} \and
Jieqiong Zhang\inst{1} \and 
Tao Du \inst{1,2,3} \and 
Li Yi \inst{1,2,3}
}

\authorrunning{X.~Liu et al.}

\institute{$^1$~Tsinghua University~~ $^2$~Shanghai AI Laboratory~
$^3$~Shanghai Qi Zhi Institute\\
\url{https://meowuu7.github.io/QuasiSim}}

\maketitle


\begin{figure}[htbp]
    \centering
  \includegraphics[width=0.9\textwidth]{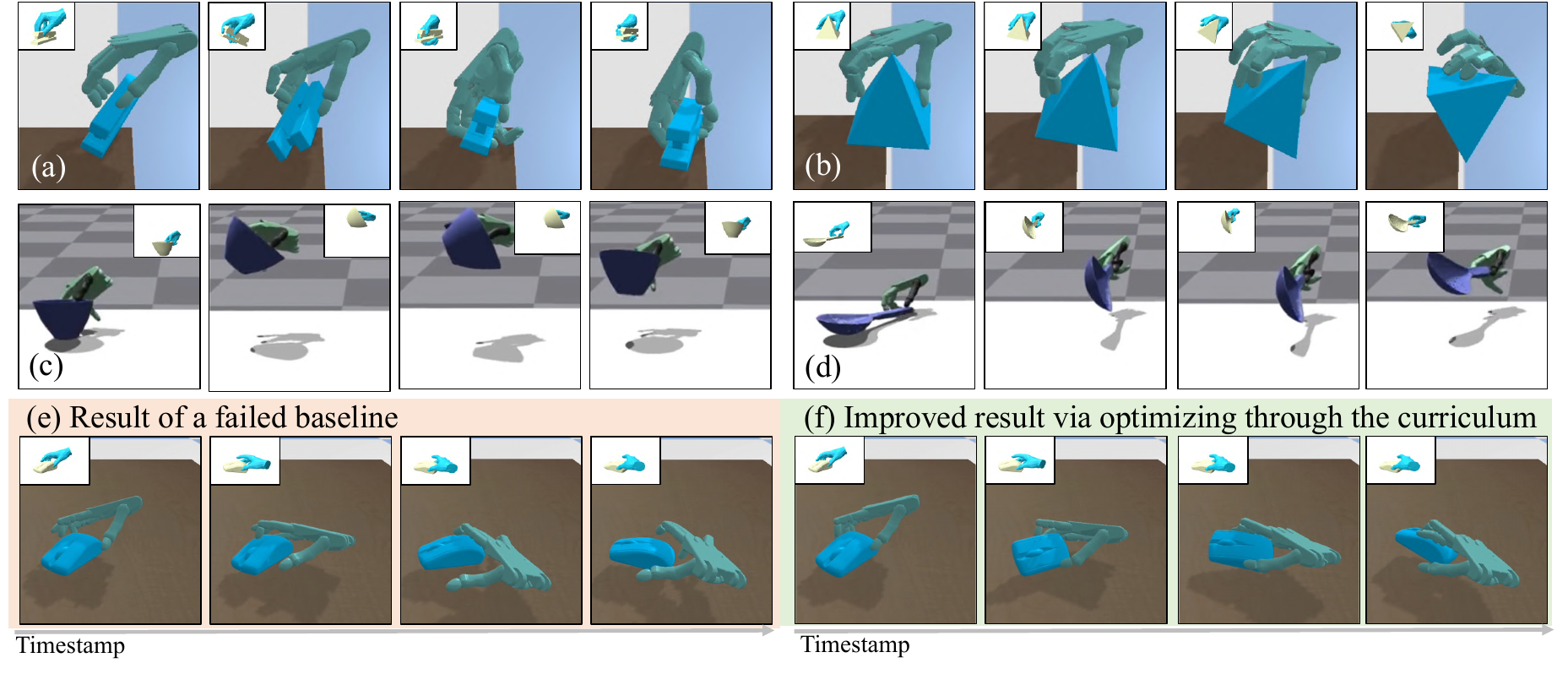}
  \vspace{-12pt}
  \caption{
  By optimizing through a \textbf{quasi-physical simulator curriculum}, we successfully transfer human demonstrations to dexterous robot hand simulations. We enable accurate tracking of complex manipulations with changing contacts 
  (\textit{Fig. (a)}), non-trivial object motions (\textit{Fig. (b)}) and intricate tool-using  (\textit{Fig. (c,d)}). Besides, our physics curriculum can substantially improve a failed baseline (\textit{Fig. (e,f)}). 
  }
  \vspace{-12pt}
  \label{fig_intro_teaser}
\end{figure}

\begin{abstract}
We explore the dexterous manipulation transfer problem by designing simulators. The task wishes to transfer human manipulations to dexterous robot hand simulations and is inherently difficult due to its intricate, highly-constrained, and discontinuous dynamics and the need to control a dexterous hand with a DoF to accurately replicate human manipulations. Previous approaches that optimize in high-fidelity black-box simulators or a modified one with relaxed constraints only demonstrate limited capabilities or are restricted by insufficient simulation fidelity. We introduce \textbf{parameterized quasi-physical simulators} and a \textbf{physics curriculum} to overcome these limitations. The key ideas are 1) balancing between fidelity and optimizability of the simulation via a curriculum of parameterized simulators, and 2) solving the problem in each of the simulators from the curriculum, with properties ranging from high task optimizability to high fidelity. We successfully enable a dexterous hand to track complex and diverse manipulations in high-fidelity simulated environments, boosting the success rate by 11\%+ from the best-performed baseline. The project website is available at \href{https://meowuu7.github.io/QuasiSim/}{QuasiSim}.
\end{abstract}

\section{Introduction}

Advancing an embodied agent's capacity to interact with the world represents a significant stride toward achieving general artificial intelligence. Due to the substantial costs and potential hazards of setting up real robots to do trial and error, the standard approach for developing embodied algorithms involves learning in physical simulators~\cite{coumans2016pybullet,makoviychuk2021isaac,todorov2012mujoco,wang2019redmax,hwangbo2018per,freeman2021brax,howell2022dojo} before transitioning to real-world deployment. In most cases, physical simulators are treated as black boxes, and extensive efforts have been devoted to developing learning and optimization methods for embodied skills within these black boxes. Despite the considerable progress~\cite{rajeswaran2017learning,chen2023visual,chen2021system,akkaya2019solving,christen2022d,zhang2023artigrasp,qin2022dexmv,liu2022herd,wu2023learning,gupta2016learning,yao2022controlvae,fussell2021supertrack,grandia2023doc,peng2018deepmimic,wang2023physhoi,mordatch2012contact}, the question like whether the simulators used are the most suitable ones is rarely discussed.
In this work, we investigate this issue and illustrate how optimizing the simulator concurrently with skill acquisition can benefit a popular yet challenging task in robot manipulation -- dexterous manipulation transfer.

The task aims at transferring human-object manipulations to a dexterous robot hand, enabling it to physically track the reference motion of both the hand and the object (see Fig.~\ref{fig_intro_teaser}). It is challenged by 1) the complex, highly constrained, non-smooth, and discontinuous dynamics with frequent contact establishment and breaking involved in the robot manipulation, 2) the requirement of precisely controlling a dexterous hand with a high DoF to densely track the manipulation at each frame, and 3) the morphology difference. Some existing works rely on high-fidelity black-box simulators, where a small difference in robot control can result in dramatically different manipulation outcomes due to abrupt contact changes, making the tracking objective highly non-smooth and hard to optimize~\cite{qin2022dexmv,chen2023visual,christen2022d,rajeswaran2017learning,andrychowicz2020learning}. In this way, their tasks are restricted to relatively simple goal-driven manipulations such as pouring and re-locating~\cite{rajeswaran2017learning,qin2022dexmv,christen2022d,zhang2023artigrasp}, in-hand re-orientation, flipping and spinning~\cite{chen2023visual,andrychowicz2020learning} with a fixed-root robot hand, or manipulating objects with simple geometry such as balls~\cite{mordatch2012contact}. 
Other approaches attempt to improve optimization by relaxing physical constraints, with a primary focus on smoothing out contact responses~\cite{howell2022trajectory,todorov2012mujoco,pang2021convex,drake,andrews2022contact}. However, their dynamics models may significantly deviate from real physics~\cite{pang2021convex}, hindering skill deployment. Consequently, we ask how to address the optimization challenge while preserving the high fidelity of the simulator.


Our key insight is that a single simulator can hardly provide both high fidelity and excellent optimizability for contact-rich dexterous manipulations.  Inspired by the line of homotopy methods~\cite{dunlavy2005homotopy,lin2023continuation,watson1989modern,liao2004homotopy}, we propose a curriculum of simulators to realize this. We start by utilizing a quasi-physical simulator to initially relax physical constraints and warm up the optimization. Subsequently, we transfer the optimization outcomes to simulators with gradually tightened physical constraints. Finally, we transition to a physically realistic simulator for skill deployment in realistic dynamics.


To realize this vision, we propose \textbf{a family of parameterized quasi-physical simulators} for contact-rich dexterous manipulation tasks. These simulators can be customized to enhance task optimizability while can also be tailored to approximate realistic physics. 
The parameterized simulator represents an articulated multi rigid body as a parameterized point set, models contact using an unconstrained parameterized spring-damper, and compensates for unmodeled effects via parameterized residual physics. Specifically, the articulated multi-body dynamics model is relaxed as the point set dynamics model. An articulated object is relaxed into a set of points, sampled from the ambient space surrounding each body's surface mesh. The resulting dynamics model combines the original articulated dynamics with the mass-point dynamics of each individual point.
Parameters are introduced to control the point set construction and the dynamics model. 
The contact model is softened as a parameterized spring-damper model~\cite{pang2021convex,andrews2022contact,geilinger2020add,marcucci2016two,suh2019comparing} with parameters introduced to control when to calculate contacts and contact spring stiffness. The residual physics network  
compensate for unmodeled effects from the analytical modeling~\cite{heiden2021neuralsim}. The parameterized simulator can be programmed for high optimizability by relaxing constraints in the analytical model and can be tailored to approximate realistic physics by learning excellent residual physics. 
We demonstrate that the challenging dexterous manipulation transfer task can be effectively addressed through curriculum optimization using a series of parameterized physical simulators.
Initially, both articulated rigid constraints and the contact model stiffness are relaxed in the simulator. 
It may not reflect physical realism but provides a good environment where the manipulation transfer problem can be solved easily. 
Subsequently, the articulated rigid constraints and the contact model are gradually tightened. Task-solving proceeds iteratively within each simulator in the curriculum.
Finally, the parameterized simulator is optimized to approximate realistic physics. Task optimization continues, yielding a dexterous hand trajectory capable of executing the manipulation in environments with realistic physics.

We demonstrate the superiority of our method and compare it with previous model-free and model-based methods on challenging manipulation sequences from three datasets, describing single-hand or bimanual manipulations with daily objects or using tools. We conduct dexterous manipulation transfer on two widely used simulators, namely Bullet~\cite{coumans2016pybullet} and Isaac Gym~\cite{makoviychuk2021isaac} to demonstrate the generality and the efficacy of our method and the capability of our quasi-physical simulator to approximate the unknown black-box physics model in the contact-rich manipulation scenario (Fig.~\ref{fig_intro_teaser}). 
We can track complex manipulations involving non-trivial object motions such as large rotations and complicated tool-using such as using a spoon to bring the water back and forth. Our approach successfully surpasses the previous best-performed method both quantitatively and qualitatively, achieving more than 11\% success rate than the previous best-performed method. Besides, optimizing through the physics curriculum can significantly enhance the performance of previously under-performed RL-based methods, almost completing the tracking problem from failure, as demonstrated in Fig.~\ref{fig_intro_teaser}. This indicates the universality of our approach to embodied AI through optimization via a physics curriculum. Thorough ablations are conducted to validate the efficacy of our designs.
%

Our contributions are three-fold:
\vspace{-7pt}
\begin{itemize}
    \item We introduce a family of parameterized quasi-physical simulators that can be configured to relax various physical constraints, facilitating skill optimization, and can also be tailored to achieve high simulation fidelity.
    \item We present a quasi-physics curriculum along with a corresponding optimization method to address the challenging dexterous manipulation transfer problem. 
    \item Extensive experiments demonstrate the effectiveness of our method in transferring complex manipulations, including non-trivial object motions and changing contacts, to a dexterous robot hand in simulation.
\end{itemize}

\vspace{-12pt}
\section{Related Works}


%
\vspace{-7pt}
\noindent\textbf{Dexterous manipulation transfer.} Transferring human manipulations to dexterous robot-hand simulations is an important topic in robot skill 
acquisition~\cite{qin2022dexmv,zhang2023learning,liu2022herd,wu2023learning,gupta2016learning,wang2023physhoi,christen2022d,zhang2023artigrasp}. Most approaches treat the simulator as black-box physics models and try to learn skills directly from that~\cite{rajeswaran2017learning,qin2022dexmv,christen2022d,chen2023visual,andrychowicz2020learning}. However, their demonstrated capabilities are restricted to relatively simple tasks. 
Another trend of work tries to relax the physics model~\cite{pang2021convex,pang2023global} to create a better environment for task optimization. However, due to the disparity between their modeling approach and realistic physics, successful trials are typically demonstrated only in their simulators, which can hardly complete the task under physically realistic dynamics.
In this work, we introduce various parameterized analytical relaxations to improve the task optimizability while compensating for unmodeled effects via residual physics networks so the fidelity would not be sacrificed. 

\noindent\textbf{Learning for simulation.} Analytical methods can hardly approximate an extremely realistic physical world despite lots of smart and tremendous efforts made in developing numerical algorithms~\cite{li2020incremental,lan2021medial,geilinger2020add,howell2022dojo}.
Recently, data-driven approaches have attracted lots of interest for their high efficiency and strong approximation ability~\cite{heiden2021neuralsim,du2023deep,pfaff2020learning,siahkoohi2019neural,deng2023learning,xiong2023neural,pfrommer2021contactnets}. 
Special network designs are proposed to learn the contact behaviour~\cite{pfrommer2021contactnets,heiden2021neuralsim}.
We in this work propose to leverage an analytical-neural hybrid approach and carefully design network modules for approximating residual contact forces in the contact-rich manipulation scenario.

\noindent\textbf{Sim-to-Sim and Sim-to-Real transfer.} 
The field of robot manipulation continues to face challenges in the areas of Sim2Sim and Sim2Real transferability~\cite{zhao2020sim}. Considering the modeling gaps, the optimal strategy learned in a specific simulator is difficult to transfer to a different simulator or the real world. 
Therefore, many techniques for solving the problem have been proposed, including imitation learning~\cite{mandikal2022dexvip,rajeswaran2017learning,schmeckpeper2020reinforcement,radosavovic2021state,qin2022one,qin2022dexmv}, transfer learning~\cite{zhuang2020comprehensive}, distillation~\cite{rusu2015policy,traore2019continual}, residual physics~\cite{zeng2020tossingbot,gao2024sim}, and efforts on bridging the gap from the dynamics model aspect~\cite{heiden2021neuralsim,zhang2024model}. Our parameterized simulators learn residual physics involved in contact-rich robot manipulations. 
By combining an analytical base with residual networks, we showcase their ability to approximate realistic physics.

\vspace{-7pt}
\section{Method}

\vspace{-7pt}




\begin{figure*}[tbp]
  \centering
  \vspace{-10pt}
  \includegraphics[width=1.0\textwidth]{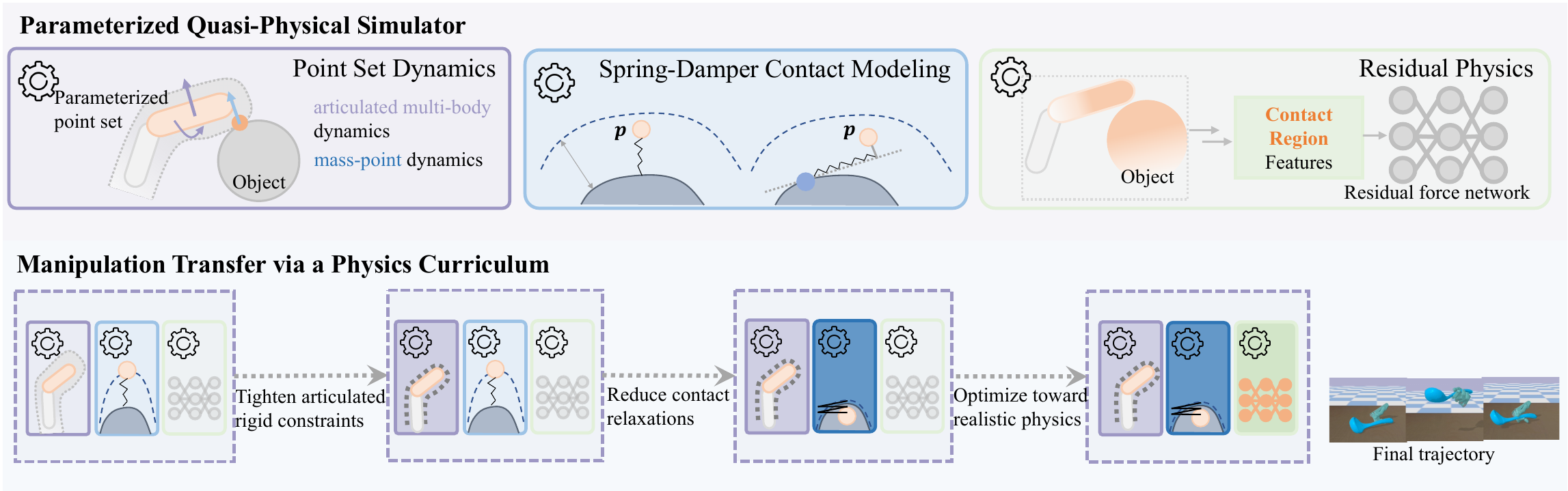}
  \vspace{-10pt}
  \vspace{-7pt}
  \caption{ 
  The \textbf{parameterized quasi-physical simulator} relaxes the articulated multi rigid body dynamics as the \textit{parameterized point set dynamics}, 
  controls the contact behavior via an unconstrained \textit{parameterized spring-damper contact model},
  and compensates for unmodeled effects via \textit{parameterized residual physics networks}. 
  We tackle the difficult dexterous manipulation transfer problem via \textbf{a physics curriculum}. 
  }
  \vspace{-10pt}
  \vspace{-7pt}
  \label{fig_method_pipeline}
\end{figure*}

Given a human manipulation demonstration, composed of a human hand mesh trajectory and an object pose trajectory $\{ \mathcal{H}, \mathcal{O} \}$, the goal is transferring the demonstration to a dexterous robot hand in simulation. Formally, we aim to optimize a control trajectory $\mathcal{A}$ that drives the dexterous hand to manipulate the object in a realistic simulated environment so that the resulting hand trajectory ${ \hat{\mathcal{H}} }$ and the object trajectory  $\mathcal{\hat{\mathcal{O}}}$ are close to the reference motion $\{ \mathcal{H}, \mathcal{O} \}$.
The problem is challenged by difficulties from the highly constrained, discontinuous, and non-smooth dynamics, the requirement of controlling a high DoF dexterous hand for tracking, and the morphology difference. 

Our method comprises two key designs to tackle the challenges: 
1) a family of parameterized quasi-physical simulators, which can be programmed to enhance the optimizability of contact-rich dexterous manipulation tasks and can also be tailored to approximate realistic physics (Section~\ref{sec:quasi_physical_sim}), and 
2)
a physics curriculum that carefully adjusts the parameters of a line of quasi-physical simulators and a strategy that solves the difficult dexterous manipulation transfer task by addressing it within each simulator in the curriculum  (Section~\ref{sec:opt_via_physics_curriculum}). 


\subsection{Parameterized Quasi-Physical Simulators } \label{sec:quasi_physical_sim}

Our quasi-physical simulator represents an articulated multi-body, \emph{i.e.,} the robotic dexterous hand, as a point set. The object is represented as a signed distance field. The base of the simulator is in an analytical form leveraging an unconstrained spring-damper contact model. Parameters are introduced to control the analytical relaxations on the articulated rigid constraints and the softness of the contact model. Additionally, neural networks are introduced to compensate for unmodeled effects beyond the analytical framework. We will elaborate on each of these design aspects below.

\begin{wrapfigure}[16]{r}{0pt}
\vspace{-7pt}
\includegraphics[width = 0.31\textwidth]{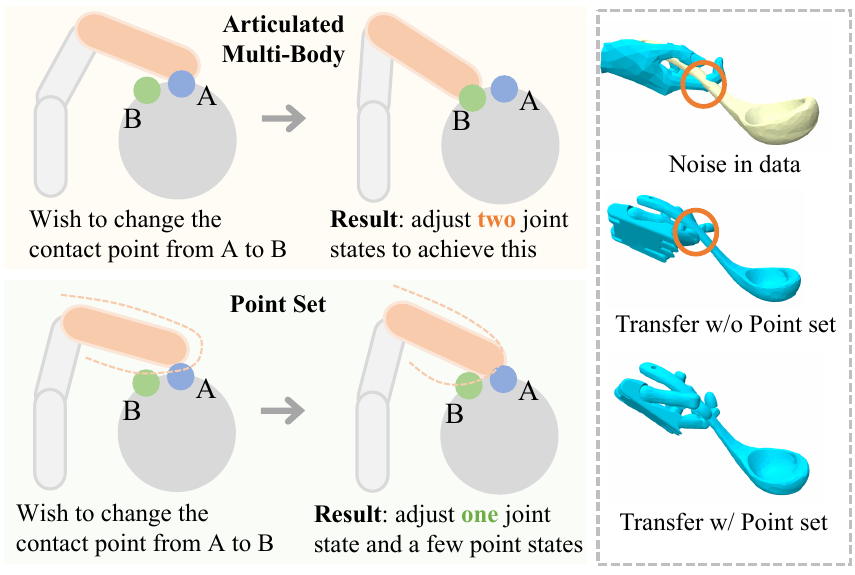}
    \caption{\textbf{Point Set} can flexibly adjust its states, avoid overfitting to data noise, and ease the difficulty brought by the morphology difference.  
    }
  \vspace{-2ex}
  \label{fig_pointdyn}
\end{wrapfigure}

\noindent\textbf{Parameterized point set dynamics.} 
Articulated multi-body represented in the reduced coordinate system~\cite{geilinger2020add,wang2019redmax} may require a large change in joint states to achieve a small adjustment in the Euclidean space. 
Moving the end effector from one point to a nearby point may require adjusting all joint states (Fig.~\ref{fig_pointdyn}). Besides, transferring the hand trajectory to a morphologically different hand requires correspondences to make the resulting trajectory close to the original one. Defining correspondences in the reduced coordinate or via sparse correspondences will make the result suffer from noise in the data, leading to unwanted results finally  (Fig.~\ref{fig_pointdyn}).
Hence, we propose relaxing an articulated multi-rigid body into a mass-point set sampled from the ambient space surrounding each body. Each point is considered attached to the body from which it is sampled and is capable of both self-actuation and actuation via joint motors. We introduce a parameter $\alpha$ to control the point set construction and the dynamics. This representation allows an articulated rigid object to behave similarly to a deformable object, providing a larger action space to adjust its state and thereby easing the control optimization problem.

Specifically, for each body of the articulated object, we sample a set of points from the ambient space near the body mesh. The point set $\mathcal{Q}$ is constructed by concatenating all sampled points together. Each point $\mathbf{p}_i \in \mathcal{Q}$ is treated as a mass point with a finite mass $\mathbf{m}_i$ and infinitesimal volume. The dynamics of the point set consist of articulated multi-body dynamics~\cite{Featherstone2007RigidBD,liu2012quick}, as well as the mass point dynamics of each point $\mathbf{p}_i$. For each $\mathbf{p}_i$, we have:
\begin{align}
{m}_i \ddot{\mathbf{x}}_i &= \mathbf{J}_{i}\mathbf{u} +  \alpha \mathbf{f}_i + \alpha  \mathbf{a}_i,
\end{align}
where $\mathbf{J}_{i}$ represents the Jacobian mapping from the generalized velocity to the point velocity $\dot{\mathbf{x}}_i$, $\mathbf{u}$ denotes the generalized joint force, $\mathbf{f}_i$ accounts for external forces acting on $\mathbf{p}_i$, and $\mathbf{a}_i\in \mathbb{R}^{3}$ represents the actuation force applied to the point $\mathbf{p}_i$. Consequently, the point set is controlled by a shared control in the reduced coordinate space $\mathbf{u}$ and per-point actuation force $\mathbf{a}_i$.

\noindent\textbf{Parameterized spring-damper contact modeling.} 
To ease the optimization challenges posed by contact-rich manipulations, which arise from contact constraints such as the non-penetration requirement and Coulomb friction law~\cite{andrews2022contact,baraff1997introduction}, as well as discontinuous dynamics involving frequent contact establishment and breaking, we propose a parameterized contact model for relaxing constraints and controlling the contact behavior. Specifically, we leverage a classical unconstrained spring-damper model~\cite{xu2021end,geilinger2020add,wang2019redmax,suh2019comparing,marcucci2016two} to model the contacts. 
This model allows us to flexibly adjust the contact behavior by tuning the contact threshold and the spring stiffness coefficients.
Intuitively, a contact model with a high threshold and low spring stiffness presents ``soft'' behaviors, resulting in a continuous and smooth optimization space.  This makes optimization through such a contact model relatively easy. Conversely, a model with a low threshold and large stiffness coefficients will produce ``stiff'' behaviors, increasing the discontinuity of the optimization space due to frequent contact establishment and breaking. However, it also becomes more physically realistic, meaning contact forces are calculated only when two objects collide, and a large force is applied to separate them if penetrations are observed, thus better satisfying the non-penetration condition.
Therefore, by adjusting the contact distance threshold and spring stiffness coefficients, we can modulate the optimizability and fidelity of the contact model.
The parameter set of the contact model comprises a distance threshold $d^c$ and spring stiffness coefficients. Next, we will delve into the details of the contact establishment, breaking, and force calculations processes.

Contacts are established between points in the manipulator's point set $\mathcal{Q}$ and the object. A point $\mathbf{p}\in \mathcal{Q}$ is considered to be in ``contact'' with the object if its signed distance to the object $\text{sd}(\mathbf{p})$ is smaller than the contact distance threshold $d^c$.
Subsequently, the object surface point nearest to $\mathbf{p}$ is identified as the corresponding contact point on the object, denoted as $\mathbf{p}^o$. The normal direction of the object point $\mathbf{p}^o$ is then determined as the contact normal direction, denoted as $\mathbf{n}^o$.
The contact force $\mathbf{f}^c$ applied from the manipulator point $\mathbf{p}$ to $\mathbf{p}^o$ is calculated as follows:
\begin{equation}
	\mathbf{f}^c = -(k^n d - k^d d\dot{d} ) \mathbf{n}^o, 
\end{equation}
where, $k^n$ represents the spring stiffness coefficient, $k^d$ denotes the damping coefficient, and $d = d^c - \text{sd}(\mathbf{p})$ is always positive. To enhance the continuity of $\mathbf{f}^c$~\cite{xu2021end}, $k^dd\dot{d}$ is used as the magnitude of the damping force, rather than $k^d \dot{d}$.


Friction forces are modeled as penalty-based spring forces~\cite{yamane2006stable,andrews2022contact}. Once a point $\mathbf{p}$ is identified as in contact with the object, with the object contact point denoted as $\mathbf{p}^o$, the contact pair is stored. Contact forces between them are continually calculated until the contact breaking conditions are met.
In more detail, the static friction force from $\mathbf{p}$ to $\mathbf{p}^o$ is calculated using a spring model: 
\begin{equation}
        \mathbf{f}_s^f = k^f \mathbf{T}_n (\mathbf{p}-\mathbf{p}^o),
\end{equation}
where $k^f$ is the friction spring stiffness coefficient, $\mathbf{T}_n = \mathbf{I} - \mathbf{n}^o{\mathbf{n}^o}^T$ is a tangential projection operator. 
When the static friction satisfies $\Vert \mathbf{f}^f_s \Vert \le \mu \Vert \mathbf{f}^c \Vert $, $\mathbf{f}_s^f$ is applied to the object point $\mathbf{p}^o$. Otherwise, the dynamic friction force is applied, and the contact breaks:
\begin{equation}
	\mathbf{f}_d^f = -\mu \Vert \mathbf{f}_s^f\Vert \frac{ \mathbf{T}_n \mathbf{v}_{\mathbf{p}\leftarrow \mathbf{p}^o}}{\Vert \mathbf{T}_n\mathbf{v}_{\mathbf{p}\leftarrow \mathbf{p}^o} \Vert },
\end{equation}
where $\mathbf{v}_{\mathbf{p}\leftarrow \mathbf{p}^o}$ is the relative velocity between $\mathbf{p}$ and $\mathbf{p}^o$.




\noindent\textbf{Parameterized residual physics.} 
The analytical designs facilitate relaxation but may limit the use of highly sophisticated and realistic dynamics models, deviating from real physics. To address this, the final component of our quasi-physical simulator is a flexible neural residual physics model~\cite{heiden2021neuralsim,ajay2018augmenting,pfrommer2021contactnets}.

Specifically, we propose to employ neural networks to learn and predict residual contact forces and friction forces based on contact-related information. For detailed residual contact force prediction, we introduce a local contact network $f_{\psi_{\text{local}}}$ that utilizes contact information identified in the parameterized contact model and predicts residual forces between each contact pair. To address discrepancies in contact region identification between the parameterized contact model and real contact region, we also incorporate a global residual network $f_{\psi_{\text{global}}}$ that predicts residual forces and torques applied directly to the object's center of mass.
In more detail, for a given contact pair $(\mathbf{p}, \mathbf{p}^o)$, the local contact network utilizes contact-related features from the local contact region, comprising geometry, per-point velocity, and per-object point normal. It then maps these features to predict the residual contact force and residual friction force between the two points in the contact pair. Additionally, the global residual network incorporates contact-related information from the global contact region, including geometry, per-point velocity, and per-object point normal, as input. It then predicts a residual force and residual torque to be applied to the object's center of mass.
Details such as contact region identification and network architectures are deferred to the Supp.
We denote the optimizable parameters in the residual physics network as $\psi = (\psi_{\text{global}}, \psi_{\text{local}})$. Through optimization of the residual physics network, we unlock the possibility of introducing highly non-linear dynamics to align our parametrized quasi-physical simulator with any realistic black-box physical simulator. 
Semi-implicit time-stepping is leveraged to make the simulation auto differentiable and easy to combine with neural networks~\cite{heiden2021neuralsim}. 


\vspace{-10pt}
\subsection{Dexterous Manipulation Transfer via a Physics Curriculum} \label{sec:opt_via_physics_curriculum}





Building upon the family of parameterized quasi-physical simulators, we present a solution to the challenging dexterous manipulation transfer problem through a physics curriculum. This curriculum consists of a sequence of parameterized simulators, ranging from those with minimal constraints and the softest contact behavior to increasingly realistic simulators. We address the problem by transferring the manipulation demonstration to the dexterous hand within each simulator across the curriculum progressively.
To elaborate further, the optimization process begins within the parameterized simulator where articulated rigid constraints are removed and the contact model is tuned to its softest level. Additionally, the residual physics networks are deactivated. This initial simulator configuration offers a friendly environment for optimization. 
Subsequently, the physics constraints are gradually tightened as we progress through each simulator within the curriculum. The task is solved iteratively within each simulator.
After reaching the most tightened analytical model, the analytical part is fixed and residual networks are activated. The simulator is gradually optimized to approximate the dynamics in a realistic physical environment. 
Concurrently, the control trajectory $\mathcal{A}$ continues to be refined in the quasi-physical simulator. Finally, we arrive at a simulator optimized to be with high fidelity and a trajectory $\mathcal{A}$ capable of guiding the dexterous hand to accurately track the demonstration within a realistically simulated physical environment.
Additionally, since object properties 
as well as system parameters 
are unknown from the kinematics-only demonstration, we set them optimizable and identify them (denoted $\mathcal{S}$) together with optimizing the hand control trajectory. 
Next we'll illustrate this in detail. 

\noindent\textbf{Transferring human demonstration via point set dynamics.}
To robustly transfer the human demonstration to a morphologically different dexterous robot hand in simulation and to overcome noise in the kinematic trajectory, we initially relax the articulated rigid constraints and transfer the kinematics human demonstration to the control trajectory of the point set. 
Specifically, the point set representation with the relaxation parameter $\alpha$ for the dynamic human hand~\cite{christen2022d} is constructed. The shared control trajectory $\mathcal{A}$ and per-point per-frame actions are optimized so that the resulting trajectory of the point set can manipulate the object according to the demonstration. After that, a point set with the same parameter $\alpha$ is constructed to represent the dexterous robot hand. 
Subsequently, the shared control trajectory $\mathcal{A}$ and per-point per-frame actions are optimized to track the manipulation accordingly. 

\noindent\textbf{Transferring through a contact model curriculum.} 
After that, the articulated rigid constraint is tightened by freezing the point set parameter $\alpha$ to zero.
The following optimization starts from a parameterized simulator with the softest contact model.
We then gradually tighten the contact model by adjusting its distance threshold, contact force spring stiffness, etc. 
By curriculum optimizing the trajectory $\mathcal{A}$ and parameters $\mathcal{S}$ in each of the quasi-physical simulators, we finally arrive at the control trajectory that can drive a dexterous hand to accomplish the tracking task in the parameterized simulator with the most tightened analytical model. 

\noindent\textbf{Optimizing towards a realistic physical environment. }
Subsequently, the residual physics network is activated and the parameterized simulator is optimized to approximate the dynamics in a realistic physical environment. We continue to optimize the hand trajectory 
in the quasi-physical simulator. Specifically, we leverage the successful trial in model-based human tracking literature~\cite{fussell2021supertrack,yao2022controlvae} and iteratively optimize the control trajectory $\mathcal{A}$ and the parameterized simulator. In more detail, the following two subproblems are iteratively solved: 
1) optimizing the quasi-physical simulator to approximate the realistic dynamics, and 2) optimizing the control trajectory $\mathcal{A}$ to complete the manipulation in the quasi-physical simulator. Gradient-based optimization is leveraged taking advantage of the differentiability of the parameterized simulator. 

After completing the optimization, 
the final control trajectory is yielded by model predictive control (MPC)~\cite{garcia1989model} based on the optimized parameterized simulator and the hand trajectory $\mathcal{A}$. Specifically, in each step, the current and the following controls in several subsequent frames are optimized to reduce the tracking error. 
More details are deferred to the Supp.

\vspace{-10pt}
\section{Experiments} \label{sec_exp}

\vspace{-10pt}


\begin{figure}[h]
  \centering
  \vspace{-10pt}
  \includegraphics[width=\textwidth]{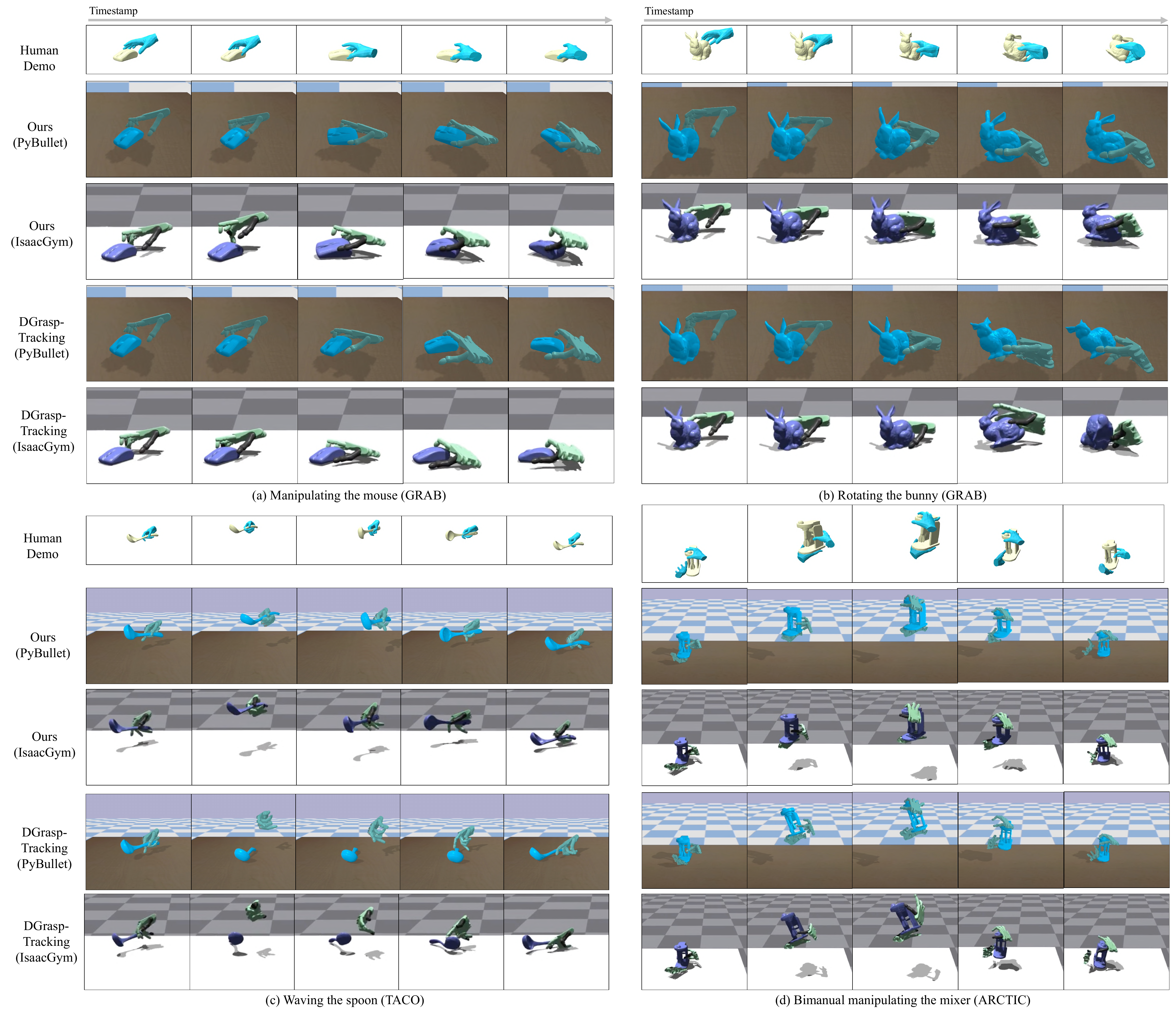}
  \vspace{-20pt}
  \caption{
  \textbf{Qualitative comparisons. }
  Please refer to \textbf{\href{https://meowuu7.github.io/QuasiSim/}{our website} and {\href{https://youtu.be/Pho3KisCsu4}{\textcolor{orange}{the accompanying video}}}} for animated results.
  }
  \label{fig_res}
  \vspace{-10pt}
\end{figure}


We conduct extensive experiments to demonstrate the effectiveness of our method. The evaluation dataset is constructed from three HOI datasets with both single-hand and bimanual manipulations (with rigid objects), with complex manipulations with non-trivial object movements, and rich and changing contacts involved (see Section~\ref{sec:exp_settings}). 
We use Shadow hand~\cite{shadowhand} and test in two simulators widely used in the embodied AI community: Bullet~\cite{coumans2016pybullet} and Isaac Gym~\cite{makoviychuk2021isaac}. 
We compare our method with both model-free approaches and model-based strategies and demonstrate the superiority of our method both quantitatively and qualitatively. We can track complex contact-rich manipulations with large object rotations, back-and-forth object movements, and changing contacts successfully in both of the two simulators, while the best-performed baseline fails (see Section~\ref{sec:retargeting_result}, Fig.~\ref{fig_res}). On average, we boost the tracking success rate by 11\%+ from the previous best-performed (see Section~\ref{sec:retargeting_result}). We make further analysis and discussions and show that the core philosophy of our work, optimizing through a quasi-physics curriculum, is potentially general and can help improve the performance of a model-free baseline (see Section~\ref{sec:exp_further_analysis}). 



\subsection{Experimental Settings} \label{sec:exp_settings}

\noindent\textbf{Datasets.} 
Our evaluation dataset is compiled from three distinct sources, namely GRAB~\cite{taheri2020grab}, containing single-hand interactions with daily objects, TACO~\cite{liu2024taco}, containing humans manipulating tools, and ARCTIC~\cite{fan2023arctic} with bimanual manipulations. 
For GRAB, we randomly sample a manipulation trajectory for each object. 
If its manipulation is extremely simple, we additionally sample one trajectory for it. 
The object is not considered if its corresponding manipulation is bimanual such as \texttt{binoculars}, involves other body parts such as \texttt{bowl}, or with detailed part movements such as the \texttt{game} \texttt{controller}.
The number of manipulation sequences from GRAB is 27. 
For TACO~\cite{liu2024taco}, we acquire data by contacting authors. 
We randomly select one sequence for each right-hand tool object. Sequences with very low quality like erroneous object motions
are excluded. 14 trajectories in total are selected finally. 
For ARCTIC~\cite{fan2023arctic}, we randomly select one sequence for each object from its available manipulation trajectories, resulting in 10 sequences in total. More details are deferred to the Supp. 

\noindent\textbf{Metrics.} 
We introduce three distinct metrics to assess the quality of object tracking, the accuracy of hand tracking, and the overall success of the tracking task: 
1) Per-frame average object rotation error: $R_{\text{err}} = \frac{1}{N} \sum_{n=1}^N (1 - (\mathbf{q}_n \cdot \hat{\mathbf{q}}_n))$, where $\mathbf{q}_n$ is the ground-truth orientation and $\hat{\mathbf{q}}_n$ is the tracked result, represented in quaternion. 2) Per-frame average object translation error: $T_{\text{err}} = \frac{1}{N} \sum_{n=1}^N \Vert \mathbf{t}_n - \hat{\mathbf{t}}_n \Vert$, where $\mathbf{t}$ and $\mathbf{t}_n$ are ground-truth and tracked translations respectively. 3) Mean Per-Joint Position Error ($\text{MPJPE}$) $=  \frac{1}{N} \sum_{n=1}^N  \Vert \mathbf{J}_n - \hat{\mathbf{J}}_n \Vert$~\cite{qin2023anyteleop,grandia2023doc,villegas2021contact}, where $\mathbf{J}_n$ and $\hat{\mathbf{J}}_n$ are keypoints of GT human hand and the simulated robot hand respectively. 
We manually define the keypoints and the correspondences to the human hand keypoints for the Shadow hand. 
4) Per-frame average hand Chamfer Distance: $\text{CD} = \frac{1}{N} \sum_{n=1}^N \text{Chamfer-Distance} (\mathbf{H}_n - \hat{\mathbf{H}}_n )$, for evaluating whether the Shadow hand can ``densely'' track the demonstration. 5) Success rate: a tracking is regarded as successful if the object rotation error $R_{\text{err}}$, object translation error $T_{\text{err}}$, and the hand tracking error $\text{MPJPE}$ 
are smaller than their corresponding threshold. Three success rates are calculated using three different thresholds, namely $10^\circ-10cm-10cm$, $15^\circ-15cm-15cm$.

\noindent\textbf{Baselines.}
We compare with two trends of baselines. 
For model-free approaches, since there is no prior work with exactly the same problem setting as us, we try to modify and improve a goal-driven rigid object manipulation method DGrasp~\cite{christen2022d} into two methods for tracking: 1) DGrasp-Base, where the method is almost kept with same with the original DGrasp. We use the first frame where the hand and the object are in contact with each other as the reference frame. Then the policy is trained to grasp the object according to the reference hand and object goal at first. After that, only the root is guided to complete the task. 2) DGrasp-Tracking, where we divide the whole sequence into several subsequences, each of which has 10 frames, and define the end frame of the subsequence as the reference frame. Then the grasping policy is used to guide the hand and gradually track the object according to the hand and the object pose of each reference frame. 
We improve the DGrasp-Tracking by optimizing the policy through the quasi-physical curriculum and creating ``DGrasp-Tracking (w/ Curriculum)'' trying to improve its performance. 
For model-based methods, we compare with Control-VAE~\cite{yao2022controlvae} and traditional MPC approaches. 
For Control-VAE, we modify its implementation for the manipulation tracking task. We additionally consider three differentiable physics models to conduct model-predictive control for solving the task. 
Taking the analytical model with the most tightened contact model 
as the base model (``MPC (w/ base sim.)''), we further augment it with a general state-of-the-art contact smoothing for robot manipulation~\cite{suh2022bundled} and create ``MPC (w/ base sim. w/ soften)''.
Details of baseline models are deferred to the Supp.

\noindent\textbf{Training and evaluation settings.}
The physics curriculum is composed of three stages. In the first stage, the parameter $\alpha$ varies from 0.1 to 0.0 and the contact model stiffness is relaxed to the softest level. 
In the second stage, $\alpha$ is fixed and the contact model stiffness varies from the softest version to the most tightened level gradually through eight stages. Details w.r.t. parameter settings are deferred to the Supp. 
In the first two stages, we alternately optimize the trajectory $\mathcal{A}$ and parameters $\mathcal{S}$. In each optimization iteration, the $\mathcal{A}$ is optimized for 100 steps while $\mathcal{S}$ is optimized for 1000 steps. In the third stage,  $\mathcal{A}$ and $\psi$ are optimized for 256 steps in each iteration. For time-stepping, $\text{d}t$ is set to $5\times10^{-4}$ in the parameterized and the target simulators. 
The articulated multi-body is controlled by joint motors and root velocities in the parameterized quasi-physical simulator while PD control~\cite{tan2012advances}  is leveraged in the target simulators.


\vspace{-10pt}

\subsection{Dexterous Manipulating Tracking}  \label{sec:retargeting_result}

\vspace{-6pt}

We conducted thorough experiments in two widely used simulators~\cite{coumans2016pybullet,makoviychuk2021isaac}. We treat them as realistic simulated physical environments with high fidelity and wish to track the manipulation in them. In summary, we can control a dexterous hand to complete a wide range of the manipulation tracking tasks with non-trivial object movements and changing contacts. As presented in Table~\ref{tb_exp_main}, we can achieve significantly higher success rates calculated under three thresholds than the best-performed baseline in both tested simulators. Fig.~\ref{fig_res} showcases qualitative examples and comparisons.  Please check out \href{https://meowuu7.github.io/QuasiSim/}{our website} and \href{https://youtu.be/Pho3KisCsu4}{video} for animated results. 
\begin{table*}[t]
    \centering
    \caption{ 
    \textbf{Quantitative evaluations and comparisons to baselines.}  \bred{Bold red} numbers for best values and \iblue{italic blue} values for the second best-performed ones. 
    } 
    \vspace{-10pt}
    \resizebox{1.0\textwidth}{!}{%
\begin{tabular}{@{\;}lllccccc@{\;}}
        \toprule
        
        Simulator & ~ & Method & \makecell[c]{$R_{\text{err}}$ ($^\circ, \downarrow$)} & \makecell[c]{$T_{\text{err}}$ (${cm}, \downarrow$)}  &  MPJPE (${mm}, \downarrow$)  & CD (${mm}, \downarrow$) & Success Rate ($\%, \uparrow$)   \\

        \cmidrule(l{0pt}r{1pt}){1-3}
        \cmidrule(l{2pt}r{2pt}){4-8}

        \multirow{7}{*}{ Bullet } & \multirow{3}{*}{ \makecell[c]{Model \\ Free} } & DGrasp-Base & 44.24 & 5.82 & 40.55 & 16.37 &  0/13.73/15.69
        \\ 
        ~ & ~ & DGrasp-Tracking & 44.45 & 5.04 & 37.56 & 14.72 &  0/15.69/15.69
        \\ 
        ~ & ~ & \makecell[l]{DGrasp-Tracking (w/ curric.)} & 33.86 & 4.60 & 30.47 & 13.53 &  7.84/\iblue{23.53}/\iblue{37.25} 
        \\ 
        \cmidrule(l{1pt}r{1pt}){2-3}

        ~ & \multirow{2}{*}{ \makecell[c]{Model \\ Based} } & Control-VAE & 42.45 & \iblue{2.73} & 25.21 & 10.94 &  0/15.68/23.53
        \\ 
        ~ & ~ & MPC (w/ base sim.) & 32.56 & 3.67 & \iblue{24.62} & \iblue{10.80} & 0/15.68/31.37
        \\ 
        ~ & ~ &  \makecell[l]{MPC (w/ base sim. w/ soften)} & \iblue{31.89} & 3.63 & 28.26 & 11.31 & 0/21.57/\iblue{37.25}
        \\ 
        
        \cmidrule(l{1pt}r{1pt}){2-3}
        \cmidrule(l{1pt}r{1pt}){4-8}

        ~ & ~ & Ours & \bred{24.21} & \bred{1.97} & \bred{24.40} & \bred{9.85} & \bred{27.45}/\bred{37.25}/\bred{58.82}
        \\ 
        
        \cmidrule(l{1pt}r{1pt}){1-3}
        \cmidrule(l{1pt}r{1pt}){4-8}

        \multirow{7}{*}{ Isaac Gym } & \multirow{3}{*}{ \makecell[c]{Model \\ Free} } & DGrasp-Base & 36.41 & 4.56 & 50.97 & 18.78 & 0/7.84 /7.84 
        \\ 
        ~ & ~ & DGrasp-Tracking & 44.71 & 5.57 & 41.53 & 16.72 & 0/0/7.84 
        \\ 
        ~ & ~ & \makecell[l]{DGrasp-Tracking (w/ curric,)} & 38.75 & 5.13 & 40.09 & 16.26 & 0/\iblue{23.53}/\iblue{31.37} 
        \\ 
        \cmidrule(l{1pt}r{1pt}){2-3}

        ~ & \multirow{3}{*}{ \makecell[c]{Model \\ Based} } & Control-VAE & \iblue{35.40} & 4.61 & 27.63 & 13.17 & 0/13.73/29.41
        \\ 
        ~ & ~ & MPC (w/ base sim.) & 37.23 & 4.73 & \bred{23.19} & \bred{9.75} & 0/15.69/\iblue{31.37}
        \\ 
        ~ & ~ &  \makecell[l]{MPC (w/ base sim. w/ soften)} & 36.40 & \iblue{4.46} & \iblue{23.27} & 10.34 & 0/9.80/23.53
        \\ 
        
        \cmidrule(l{1pt}r{1pt}){2-3}
        \cmidrule(l{1pt}r{1pt}){4-8}

        ~ & ~ & Ours & \bred{25.97} & \bred{2.08} & {25.33} & \iblue{10.31} & \bred{21.57}/\bred{43.14}/\bred{56.86}
        \\ 

        \bottomrule
 
    \end{tabular}
    }
    \vspace{-10pt}
    \label{tb_exp_main}
\end{table*}

\noindent\textbf{Complex manipulations.} 
For examples shown in Fig.~\ref{fig_res}, we can complete the tracking task on examples with large object re-orientations and complicated tool-using (\textit{Fig. (a,b,c)}). However, DGrasp-Tracking fails to establish sufficient contact for correctly manipulating the object. In more detail, in Fig.~\ref{fig_res}(b), the bunny gradually bounced out from its hand in Bullet, while our method does not suffer from this difficulty. In Fig.~\ref{fig_res}(c), the spoon can be successfully picked up and waved back-and-forth in our method, while DGrasp-Tracking loses the track right from the start. 

\noindent\textbf{Bimanual manipulations.}   
We are also capable of tracking bimanual manipulations. As shown in the example in Fig.~\ref{fig_res}(d), where two hands collaborate to relocate the object, DGrasp-Tracking fails to accurately track the object, while our method significantly outperforms it.
 


\vspace{-15pt}
\subsection{Further Analysis and Discussions} \label{sec:exp_further_analysis}
\begin{figure}[htbp]
\vspace{-20pt}
  \centering
  \includegraphics[width=\textwidth]{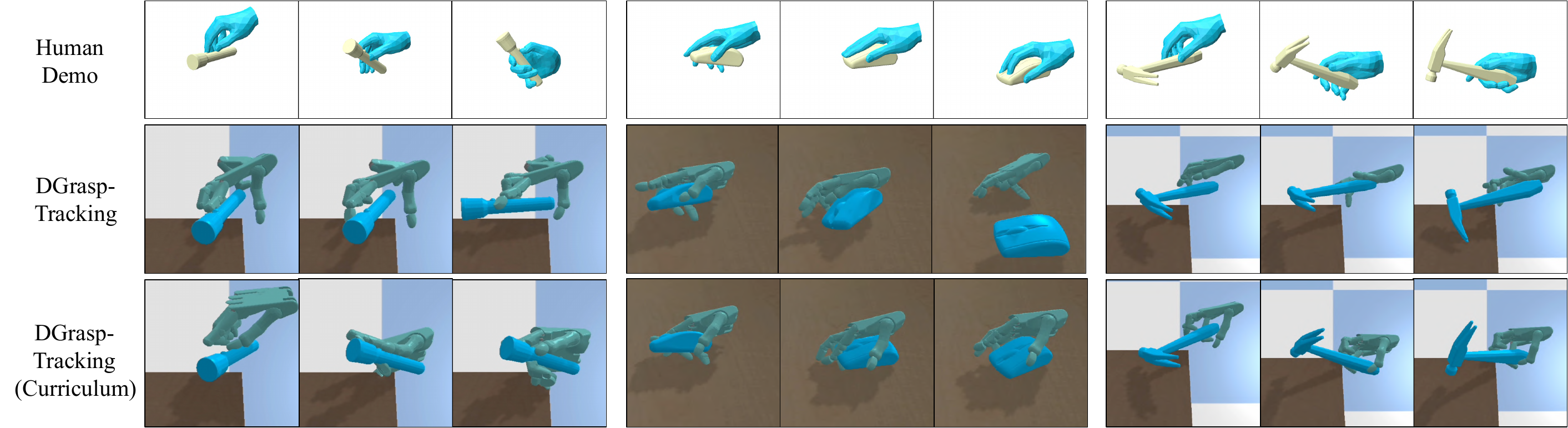}
  \caption{
  Visual evidence on boosting DGrasp-Tracking's performance via optimizing it through a physics curriculum. 
  }
  \label{fig_rl_curriculum}
  \vspace{-16pt}
\end{figure}

\noindent\textbf{Could model-free methods benefit from the physics curriculum?}
In addition to the demonstrated merits of our quasi-physical simulators, we further explore whether model-free strategies can benefit from them. We introduce the ``DGrasp-Tracking (w/ Curriculum)'' method and compare its performance with the original DGrasp-Tracking model. As shown in Table~\ref{tb_exp_main} and the visual comparisons in Fig.~\ref{fig_rl_curriculum}, the DGrasp-Tracking model indeed benefits from a well-designed physics curriculum. For example, as illustrated in Fig.~\ref{fig_rl_curriculum}, the curriculum can significantly improve its performance, enabling it to nearly complete challenging tracking tasks where the original version struggles.

\vspace{-10pt}
\section{Ablation Study} \label{sec:ablations}

\vspace{-7pt}

\begin{figure}[htbp]
  \centering
  \vspace{-10pt}
  \includegraphics[width=0.7\textwidth]{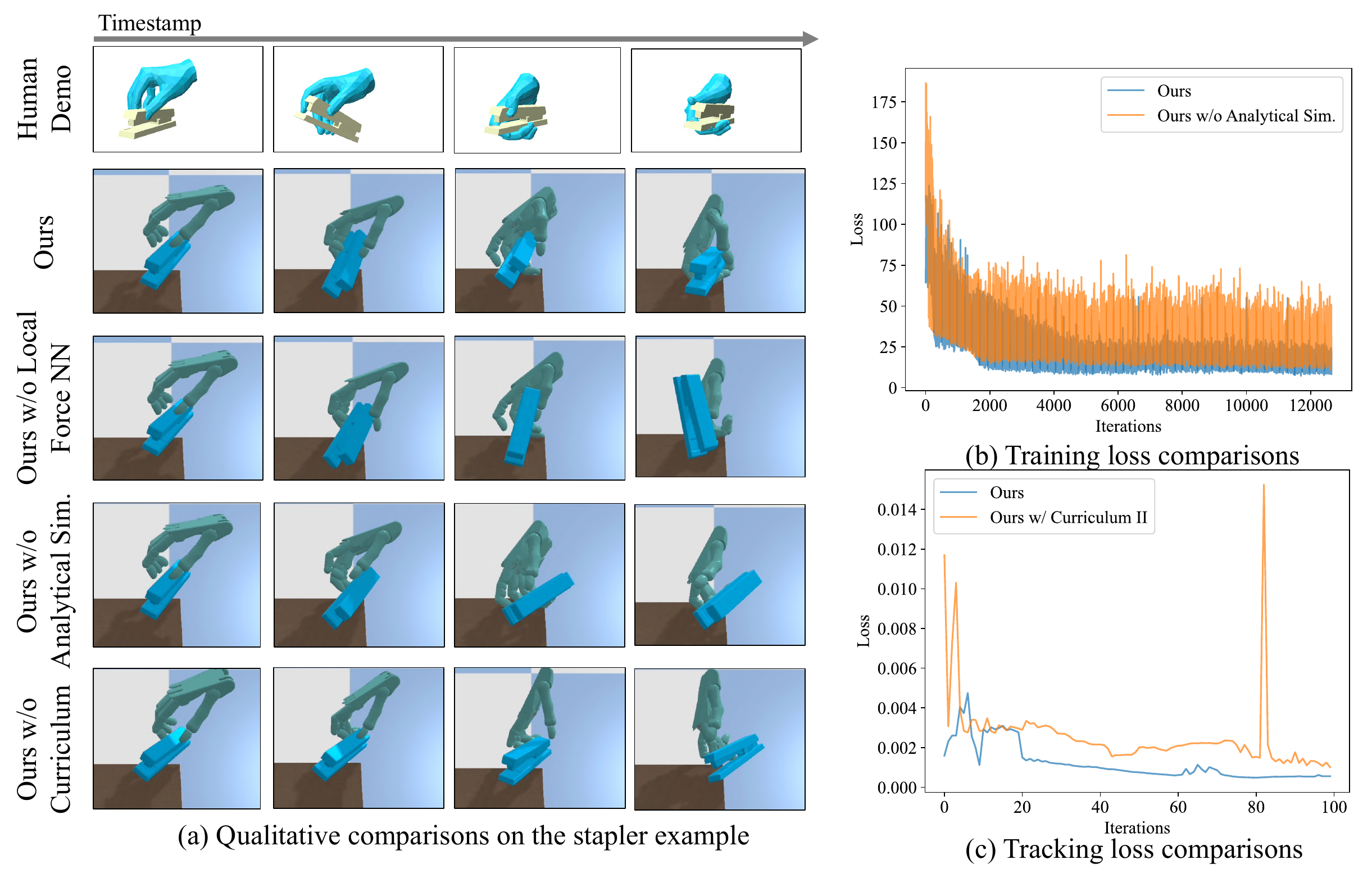}
  \vspace{-13pt}
  \caption{
  (a) Qualitative comparisons between our full method and the ablated models; (b) Training loss curve comparisons; 
  (c) Tracking loss curve comparisons.
  }
  \vspace{-20pt}
  \label{fig_abl_stapler}
\end{figure}

\begin{table*}[t]
    \centering
    \caption{ 
    \textbf{Ablation studies.}  \bred{Bold red} numbers for best values and \iblue{italic blue} values for the second best-performed ones. The simulation environment is Bullet. 
    } 
    \vspace{-10pt}
    \resizebox{\textwidth}{!}{%
\begin{tabular}{@{\;}lccccc@{\;}}
        \toprule

        Method & \makecell[c]{$R_{\text{err}}$ ($^\circ, \downarrow$)} & \makecell[c]{$T_{\text{err}}$ (${cm}, \downarrow$)}  &  MPJPE (${mm}, \downarrow$)  & CD (${mm}, \downarrow$) & Success Rate ($\%, \uparrow$)    \\

        \cmidrule(l{0pt}r{1pt}){1-1}
        \cmidrule(l{2pt}r{2pt}){2-6}

        \makecell[l]{Ours w/o Analytical Sim.} & 44.27 & 4.39 & 29.84 & 12.91 &  0/13.73/25.49
        \\ 
        \makecell[l]{Ours w/o Residual Physics}  & 33.69 & 3.81 & \iblue{26.57} & 10.34 & 5.88/23.53/41.18
        \\ 
        \makecell[l]{Ours w/o Local Force NN} & 35.98 & 2.90 & 32.87 & 12.44 &  0/19.61/35.29
        \\ 
        \makecell[l]{Ours w/o Curriculum} & 42.40 & 4.87 & 32.61 & 13.37 &  0/17.64/29.41
        \\ 
        \makecell[l]{Ours w/ Curriculum II} & \iblue{29.58} & \iblue{2.33} & 31.61 & \iblue{10.29} &  \iblue{11.76}/\iblue{27.45}/\iblue{50.98}
        \\ 
        
        Ours & \bred{24.21} & \bred{1.97} & \bred{24.40} & \bred{9.85} & \bred{27.45}/\bred{37.25}/\bred{58.82}
        \\ 
        \bottomrule
 
    \end{tabular}
    }
    \vspace{-20pt}
    \label{tb_exp_ablations}
\end{table*}

We conduct a wide range of ablation studies to validate the effectiveness of some of our crucial designs, including the parameterized analytical physics model, the parameterized residual physics, the role of the local force network, the necessity of introducing a physics curriculum into the optimization, and how the design on the curriculum stages affects the result. 


\noindent\textbf{Parameterized analytical model.}
The skeleton of the quasi-physical simulator is an analytical physics model. The intuition is that the parameterized simulator with such physical bias can be optimized towards a realistic simulator more easily than training pure neural networks for approximating. To validate this, we ablate the analytical model and use neural networks to approximate physics in Bullet directly (denoted as ``Ours w/o Analytical Sim.''). The quantitative (Table~\ref{tb_exp_ablations}) and qualitative (Fig.~\ref{fig_abl_stapler}) results indicate that the physical biases brought by the analytical model could help the parameterized simulator to learn better physics in the contact-rich scenario. For instance, in the example demonstrated in Fig.~\ref{fig_abl_stapler}, the ablated version fails to guide the robot hand to successfully pinch the object in the second figure. 

\noindent\textbf{Parameterized residual physics.}
To validate the necessity of introducing residual force networks to close the gap between the physics modeled in the parameterized analytical simulator and that of a realistic simulator, we ablate the parameterized force network and create a version named ``Ours w/o Residual Physics''. Table~\ref{tb_exp_ablations} demonstrated its role in enabling the parameterized simulator to approximate realistic physics models.

\noindent\textbf{Local residual force network.} 
To adequately leverage state and contact-related information for predicting residual contact forces, we propose to use two types of networks: 1) a local force network for per contact pair residual forces and 2) a global network for additionally compensating. 
The local network is introduced for fine-grained approximation. We ablate this design and compare the result with our full model to validate this  (see Fig.~\ref{fig_abl_stapler} and Table~\ref{tb_exp_main}). 


\noindent\textbf{Optimizing through an analytical physics curriculum.}
We further investigate the effectiveness of the analytical curriculum design and how its design influences the result. Specifically, we create two ablated versions: 
1) ``Ours w/o Curriculum'', where the optimization starts directly from the parameterized analytical model with articulated rigid constraints tightened and the stiffest contact model, and 2) ``Ours w/ Curriculum II'', where we move some stages out from the original curriculum. 
Table~\ref{tb_exp_ablations}  and Fig.~\ref{fig_abl_stapler} demonstrate that both the curriculum and the optimization path will affect the model's performance. 

\vspace{-17pt}

\section{Conclusion and Limitations}

\vspace{-10pt}
In this work, we investigate creating better simulators for solving complex robotic tasks involving complicated dynamics where the previous best-performed optimization strategy fails. We present a family of parameterized quasi-physical simulators that can be both programmed to relax various constraints for task optimization and can be tailored to 
approximate realistic physics. 
We tackle the difficult manipulation transfer task via a physics curriculum.

\noindent\textbf{Limitations.} The method is limited by the relatively simple spring-damper model for contact constraint relaxation. Introducing delicate analytical contact models to parameterized simulators is an interesting research direction. 

\section*{Acknowledgement}
We thank Prof. Rui Chen, Yixuan Guan, and Taoran Jiang for their valuable support in providing the Allegro hardware and setting up the environment setup during the rebuttal period.


\bibliographystyle{splncs04}
\bibliography{main}

\begin{thebibliography}{10}
\providecommand{\url}[1]{\texttt{#1}}
\providecommand{\urlprefix}{URL }
\providecommand{\doi}[1]{https://doi.org/#1}

\bibitem{ajay2018augmenting}
Ajay, A., Wu, J., Fazeli, N., Bauza, M., Kaelbling, L.P., Tenenbaum, J.B., Rodriguez, A.: Augmenting physical simulators with stochastic neural networks: Case study of planar pushing and bouncing. In: 2018 IEEE/RSJ International Conference on Intelligent Robots and Systems (IROS). pp. 3066--3073. IEEE (2018)

\bibitem{akkaya2019solving}
Akkaya, I., Andrychowicz, M., Chociej, M., Litwin, M., McGrew, B., Petron, A., Paino, A., Plappert, M., Powell, G., Ribas, R., et~al.: Solving rubik's cube with a robot hand. arXiv preprint arXiv:1910.07113  (2019)

\bibitem{andrews2022contact}
Andrews, S., Erleben, K., Ferguson, Z.: Contact and friction simulation for computer graphics. In: ACM SIGGRAPH 2022 Courses, pp. 1--172 (2022)

\bibitem{andrychowicz2020learning}
Andrychowicz, O.M., Baker, B., Chociej, M., Jozefowicz, R., McGrew, B., Pachocki, J., Petron, A., Plappert, M., Powell, G., Ray, A., et~al.: Learning dexterous in-hand manipulation. The International Journal of Robotics Research  \textbf{39}(1),  3--20 (2020)

\bibitem{arunachalam2023dexterous}
Arunachalam, S.P., Silwal, S., Evans, B., Pinto, L.: Dexterous imitation made easy: A learning-based framework for efficient dexterous manipulation. In: 2023 ieee international conference on robotics and automation (icra). pp. 5954--5961. IEEE (2023)

\bibitem{bahl2023affordances}
Bahl, S., Mendonca, R., Chen, L., Jain, U., Pathak, D.: Affordances from human videos as a versatile representation for robotics. In: Proceedings of the IEEE/CVF Conference on Computer Vision and Pattern Recognition. pp. 13778--13790 (2023)

\bibitem{baraff1997introduction}
Baraff, D.: An introduction to physically based modeling: rigid body simulation ii—nonpenetration constraints. SIGGRAPH course notes pp. D31--D68 (1997)

\bibitem{bharadhwaj2023towards}
Bharadhwaj, H., Gupta, A., Kumar, V., Tulsiani, S.: Towards generalizable zero-shot manipulation via translating human interaction plans. arXiv preprint arXiv:2312.00775  (2023)

\bibitem{chen2023human}
Chen, G., Cui, T., Zhou, T., Peng, Z., Hu, M., Wang, M., Yang, Y., Yue, Y.: Human demonstrations are generalizable knowledge for robots. arXiv preprint arXiv:2312.02419  (2023)

\bibitem{chen2023visual}
Chen, T., Tippur, M., Wu, S., Kumar, V., Adelson, E., Agrawal, P.: Visual dexterity: In-hand reorientation of novel and complex object shapes. Science Robotics  \textbf{8}(84),  eadc9244 (2023). \doi{10.1126/scirobotics.adc9244}, \url{https://www.science.org/doi/abs/10.1126/scirobotics.adc9244}

\bibitem{chen2021system}
Chen, T., Xu, J., Agrawal, P.: A system for general in-hand object re-orientation. Conference on Robot Learning  (2021)

\bibitem{christen2022d}
Christen, S., Kocabas, M., Aksan, E., Hwangbo, J., Song, J., Hilliges, O.: D-grasp: Physically plausible dynamic grasp synthesis for hand-object interactions. In: Proceedings of the IEEE/CVF Conference on Computer Vision and Pattern Recognition. pp. 20577--20586 (2022)

\bibitem{coumans2016pybullet}
Coumans, E., Bai, Y.: Pybullet, a python module for physics simulation for games, robotics and machine learning  (2016)

\bibitem{deng2023learning}
Deng, Y., Yu, H.X., Wu, J., Zhu, B.: Learning vortex dynamics for fluid inference and prediction. arXiv preprint arXiv:2301.11494  (2023)

\bibitem{du2023deep}
Du, T.: Deep learning for physics simulation. In: ACM SIGGRAPH 2023 Courses, pp. 1--25 (2023)

\bibitem{dunlavy2005homotopy}
Dunlavy, D.M., O'Leary, D.P.: Homotopy optimization methods for global optimization. Tech. rep., Sandia National Laboratories (SNL), Albuquerque, NM, and Livermore, CA~… (2005)

\bibitem{fan2023arctic}
Fan, Z., Taheri, O., Tzionas, D., Kocabas, M., Kaufmann, M., Black, M.J., Hilliges, O.: {ARCTIC}: A dataset for dexterous bimanual hand-object manipulation. In: Proceedings IEEE Conference on Computer Vision and Pattern Recognition (CVPR) (2023)

\bibitem{Featherstone2007RigidBD}
Featherstone, R.: Rigid body dynamics algorithms (2007), \url{https://api.semanticscholar.org/CorpusID:58437819}

\bibitem{freeman2021brax}
Freeman, C.D., Frey, E., Raichuk, A., Girgin, S., Mordatch, I., Bachem, O.: Brax--a differentiable physics engine for large scale rigid body simulation. arXiv preprint arXiv:2106.13281  (2021)

\bibitem{fussell2021supertrack}
Fussell, L., Bergamin, K., Holden, D.: Supertrack: Motion tracking for physically simulated characters using supervised learning. ACM Transactions on Graphics (TOG)  \textbf{40}(6),  1--13 (2021)

\bibitem{gao2024sim}
Gao, J., Michelis, M.Y., Spielberg, A., Katzschmann, R.K.: Sim-to-real of soft robots with learned residual physics. arXiv preprint arXiv:2402.01086  (2024)

\bibitem{garcia1989model}
Garcia, C.E., Prett, D.M., Morari, M.: Model predictive control: Theory and practice—a survey. Automatica  \textbf{25}(3),  335--348 (1989)

\bibitem{geilinger2020add}
Geilinger, M., Hahn, D., Zehnder, J., B{\"a}cher, M., Thomaszewski, B., Coros, S.: Add: Analytically differentiable dynamics for multi-body systems with frictional contact. ACM Transactions on Graphics (TOG)  \textbf{39}(6),  1--15 (2020)

\bibitem{grandia2023doc}
Grandia, R., Farshidian, F., Knoop, E., Schumacher, C., Hutter, M., B{\"a}cher, M.: Doc: Differentiable optimal control for retargeting motions onto legged robots. ACM Transactions on Graphics (TOG)  \textbf{42}(4),  1--14 (2023)

\bibitem{guo2023learning}
Guo, D.: Learning multi-step manipulation tasks from a single human demonstration. arXiv preprint arXiv:2312.15346  (2023)

\bibitem{gupta2016learning}
Gupta, A., Eppner, C., Levine, S., Abbeel, P.: Learning dexterous manipulation for a soft robotic hand from human demonstrations. In: 2016 IEEE/RSJ International Conference on Intelligent Robots and Systems (IROS). pp. 3786--3793. IEEE (2016)

\bibitem{heiden2021neuralsim}
Heiden, E., Millard, D., Coumans, E., Sheng, Y., Sukhatme, G.S.: Neuralsim: Augmenting differentiable simulators with neural networks. In: 2021 IEEE International Conference on Robotics and Automation (ICRA). pp. 9474--9481. IEEE (2021)

\bibitem{howell2022dojo}
Howell, T.A., Le~Cleac’h, S., Kolter, J.Z., Schwager, M., Manchester, Z.: Dojo: A differentiable simulator for robotics. arXiv preprint arXiv:2203.00806  \textbf{9} (2022)

\bibitem{howell2022trajectory}
Howell, T.A., Le~Cleac’h, S., Singh, S., Florence, P., Manchester, Z., Sindhwani, V.: Trajectory optimization with optimization-based dynamics. IEEE Robotics and Automation Letters  \textbf{7}(3),  6750--6757 (2022)

\bibitem{hwangbo2018per}
Hwangbo, J., Lee, J., Hutter, M.: Per-contact iteration method for solving contact dynamics. IEEE Robotics and Automation Letters  \textbf{3}(2),  895--902 (2018)

\bibitem{lan2021medial}
Lan, L., Yang, Y., Kaufman, D., Yao, J., Li, M., Jiang, C.: Medial ipc: accelerated incremental potential contact with medial elastics. ACM Transactions on Graphics  \textbf{40}(4) (2021)

\bibitem{li2020incremental}
Li, M., Ferguson, Z., Schneider, T., Langlois, T.R., Zorin, D., Panozzo, D., Jiang, C., Kaufman, D.M.: Incremental potential contact: intersection-and inversion-free, large-deformation dynamics. ACM Trans. Graph.  \textbf{39}(4), ~49 (2020)

\bibitem{liao2004homotopy}
Liao, S.: On the homotopy analysis method for nonlinear problems. Applied mathematics and computation  \textbf{147}(2),  499--513 (2004)

\bibitem{lin2023continuation}
Lin, X., Yang, Z., Zhang, X., Zhang, Q.: Continuation path learning for homotopy optimization  (2023)

\bibitem{liu2012quick}
Liu, C.K., Jain, S.: A quick tutorial on multibody dynamics. Online tutorial, June p.~7 (2012)

\bibitem{liu2022herd}
Liu, X., Pathak, D., Kitani, K.M.: Herd: Continuous human-to-robot evolution for learning from human demonstration. arXiv preprint arXiv:2212.04359  (2022)

\bibitem{liu2024taco}
Liu, Y., Yang, H., Si, X., Liu, L., Li, Z., Zhang, Y., Liu, Y., Yi, L.: Taco: Benchmarking generalizable bimanual tool-action-object understanding. arXiv preprint arXiv:2401.08399  (2024)

\bibitem{makoviychuk2021isaac}
Makoviychuk, V., Wawrzyniak, L., Guo, Y., Lu, M., Storey, K., Macklin, M., Hoeller, D., Rudin, N., Allshire, A., Handa, A., et~al.: Isaac gym: High performance gpu-based physics simulation for robot learning. arXiv preprint arXiv:2108.10470  (2021)

\bibitem{mandikal2022dexvip}
Mandikal, P., Grauman, K.: Dexvip: Learning dexterous grasping with human hand pose priors from video. In: Conference on Robot Learning. pp. 651--661. PMLR (2022)

\bibitem{marcucci2016two}
Marcucci, T., Gabiccini, M., Artoni, A.: A two-stage trajectory optimization strategy for articulated bodies with unscheduled contact sequences. IEEE Robotics and Automation Letters  \textbf{2}(1),  104--111 (2016)

\bibitem{mordatch2012contact}
Mordatch, I., Popovi{\'c}, Z., Todorov, E.: Contact-invariant optimization for hand manipulation. In: Proceedings of the ACM SIGGRAPH/Eurographics symposium on computer animation. pp. 137--144 (2012)

\bibitem{pang2023global}
Pang, T., Suh, H.T., Yang, L., Tedrake, R.: Global planning for contact-rich manipulation via local smoothing of quasi-dynamic contact models. IEEE Transactions on Robotics  (2023)

\bibitem{pang2021convex}
Pang, T., Tedrake, R.: A convex quasistatic time-stepping scheme for rigid multibody systems with contact and friction. In: 2021 IEEE International Conference on Robotics and Automation (ICRA). pp. 6614--6620. IEEE (2021)

\bibitem{peng2018deepmimic}
Peng, X.B., Abbeel, P., Levine, S., Van~de Panne, M.: Deepmimic: Example-guided deep reinforcement learning of physics-based character skills. ACM Transactions On Graphics (TOG)  \textbf{37}(4),  1--14 (2018)

\bibitem{pfaff2020learning}
Pfaff, T., Fortunato, M., Sanchez-Gonzalez, A., Battaglia, P.W.: Learning mesh-based simulation with graph networks. arXiv preprint arXiv:2010.03409  (2020)

\bibitem{pfrommer2021contactnets}
Pfrommer, S., Halm, M., Posa, M.: Contactnets: Learning discontinuous contact dynamics with smooth, implicit representations. In: Conference on Robot Learning. pp. 2279--2291. PMLR (2021)

\bibitem{qin2023dexpoint}
Qin, Y., Huang, B., Yin, Z.H., Su, H., Wang, X.: Dexpoint: Generalizable point cloud reinforcement learning for sim-to-real dexterous manipulation. In: Conference on Robot Learning. pp. 594--605. PMLR (2023)

\bibitem{qin2022one}
Qin, Y., Su, H., Wang, X.: From one hand to multiple hands: Imitation learning for dexterous manipulation from single-camera teleoperation. IEEE Robotics and Automation Letters  \textbf{7}(4),  10873--10881 (2022)

\bibitem{qin2022dexmv}
Qin, Y., Wu, Y.H., Liu, S., Jiang, H., Yang, R., Fu, Y., Wang, X.: Dexmv: Imitation learning for dexterous manipulation from human videos. In: European Conference on Computer Vision. pp. 570--587. Springer (2022)

\bibitem{qin2023anyteleop}
Qin, Y., Yang, W., Huang, B., Van~Wyk, K., Su, H., Wang, X., Chao, Y.W., Fox, D.: Anyteleop: A general vision-based dexterous robot arm-hand teleoperation system. arXiv preprint arXiv:2307.04577  (2023)

\bibitem{radosavovic2021state}
Radosavovic, I., Wang, X., Pinto, L., Malik, J.: State-only imitation learning for dexterous manipulation. In: 2021 IEEE/RSJ International Conference on Intelligent Robots and Systems (IROS). pp. 7865--7871. IEEE (2021)

\bibitem{rajeswaran2017learning}
Rajeswaran, A., Kumar, V., Gupta, A., Vezzani, G., Schulman, J., Todorov, E., Levine, S.: Learning complex dexterous manipulation with deep reinforcement learning and demonstrations. arXiv preprint arXiv:1709.10087  (2017)

\bibitem{rusu2015policy}
Rusu, A.A., Colmenarejo, S.G., Gulcehre, C., Desjardins, G., Kirkpatrick, J., Pascanu, R., Mnih, V., Kavukcuoglu, K., Hadsell, R.: Policy distillation. arXiv preprint arXiv:1511.06295  (2015)

\bibitem{schmeckpeper2020reinforcement}
Schmeckpeper, K., Rybkin, O., Daniilidis, K., Levine, S., Finn, C.: Reinforcement learning with videos: Combining offline observations with interaction. arXiv preprint arXiv:2011.06507  (2020)

\bibitem{shadowhand}
ShadowRobot: Shadowrobot dexterous hand (2005), \url{https://www.shadowrobot. com/dexterous-hand-series/}

\bibitem{shaw2023videodex}
Shaw, K., Bahl, S., Pathak, D.: Videodex: Learning dexterity from internet videos. In: Conference on Robot Learning. pp. 654--665. PMLR (2023)

\bibitem{siahkoohi2019neural}
Siahkoohi, A., Louboutin, M., Herrmann, F.J.: Neural network augmented wave-equation simulation. arXiv preprint arXiv:1910.00925  (2019)

\bibitem{suh2019comparing}
Suh, H., Wang, Y.: Comparing effectiveness of relaxation methods for warm starting trajectory optimization through soft contact  (2019)

\bibitem{suh2022bundled}
Suh, H.J.T., Pang, T., Tedrake, R.: Bundled gradients through contact via randomized smoothing. IEEE Robotics and Automation Letters  \textbf{7}(2),  4000--4007 (2022)

\bibitem{taheri2020grab}
Taheri, O., Ghorbani, N., Black, M.J., Tzionas, D.: Grab: A dataset of whole-body human grasping of objects. In: Computer Vision--ECCV 2020: 16th European Conference, Glasgow, UK, August 23--28, 2020, Proceedings, Part IV 16. pp. 581--600. Springer (2020)

\bibitem{tan2012advances}
Tan, K.K., Wang, Q.G., Hang, C.C.: Advances in PID control. Springer Science \& Business Media (2012)

\bibitem{drake}
Tedrake, R., the Drake Development~Team: Drake: Model-based design and verification for robotics (2019), \url{https://drake.mit.edu}

\bibitem{todorov2012mujoco}
Todorov, E., Erez, T., Tassa, Y.: Mujoco: A physics engine for model-based control. In: 2012 IEEE/RSJ international conference on intelligent robots and systems. pp. 5026--5033. IEEE (2012)

\bibitem{traore2019continual}
Traor{\'e}, R., Caselles-Dupr{\'e}, H., Lesort, T., Sun, T., D{\'\i}az-Rodr{\'\i}guez, N., Filliat, D.: Continual reinforcement learning deployed in real-life using policy distillation and sim2real transfer. arXiv preprint arXiv:1906.04452  (2019)

\bibitem{villegas2021contact}
Villegas, R., Ceylan, D., Hertzmann, A., Yang, J., Saito, J.: Contact-aware retargeting of skinned motion. In: Proceedings of the IEEE/CVF International Conference on Computer Vision. pp. 9720--9729 (2021)

\bibitem{mesh2sdf}
Wang, P.S.: Mesh2sdf: Converts an input mesh to a signed distance field (2022), \url{https://github.com/wang-ps/mesh2sdf}

\bibitem{wang2019redmax}
Wang, Y., Weidner, N.J., Baxter, M.A., Hwang, Y., Kaufman, D.M., Sueda, S.: Redmax: Efficient \& flexible approach for articulated dynamics. ACM Transactions on Graphics (TOG)  \textbf{38}(4),  1--10 (2019)

\bibitem{wang2023physhoi}
Wang, Y., Lin, J., Zeng, A., Luo, Z., Zhang, J., Zhang, L.: Physhoi: Physics-based imitation of dynamic human-object interaction. arXiv preprint arXiv:2312.04393  (2023)

\bibitem{watson1989modern}
Watson, L.T., Haftka, R.T.: Modern homotopy methods in optimization. Computer Methods in Applied Mechanics and Engineering  \textbf{74}(3),  289--305 (1989)

\bibitem{wu2023learning}
Wu, Y.H., Wang, J., Wang, X.: Learning generalizable dexterous manipulation from human grasp affordance. In: Conference on Robot Learning. pp. 618--629. PMLR (2023)

\bibitem{xiong2023neural}
Xiong, S., He, X., Tong, Y., Deng, Y., Zhu, B.: Neural vortex method: from finite lagrangian particles to infinite dimensional eulerian dynamics. Computers \& Fluids  \textbf{258},  105811 (2023)

\bibitem{xu2021end}
Xu, J., Chen, T., Zlokapa, L., Foshey, M., Matusik, W., Sueda, S., Agrawal, P.: An end-to-end differentiable framework for contact-aware robot design. arXiv preprint arXiv:2107.07501  (2021)

\bibitem{yamane2006stable}
Yamane, K., Nakamura, Y.: Stable penalty-based model of frictional contacts. In: Proceedings 2006 IEEE International Conference on Robotics and Automation, 2006. ICRA 2006. pp. 1904--1909. IEEE (2006)

\bibitem{yao2022controlvae}
Yao, H., Song, Z., Chen, B., Liu, L.: Controlvae: Model-based learning of generative controllers for physics-based characters. ACM Transactions on Graphics (TOG)  \textbf{41}(6),  1--16 (2022)

\bibitem{zeng2020tossingbot}
Zeng, A., Song, S., Lee, J., Rodriguez, A., Funkhouser, T.: Tossingbot: Learning to throw arbitrary objects with residual physics. IEEE Transactions on Robotics  \textbf{36}(4),  1307--1319 (2020)

\bibitem{zhang2023artigrasp}
Zhang, H., Christen, S., Fan, Z., Zheng, L., Hwangbo, J., Song, J., Hilliges, O.: Artigrasp: Physically plausible synthesis of bi-manual dexterous grasping and articulation. arXiv preprint arXiv:2309.03891  (2023)

\bibitem{zhang2024model}
Zhang, S., Liu, B., Wang, Z., Zhao, T.: Model-based reparameterization policy gradient methods: Theory and practical algorithms. Advances in Neural Information Processing Systems  \textbf{36} (2024)

\bibitem{zhang2023learning}
Zhang, Y., Clegg, A., Ha, S., Turk, G., Ye, Y.: Learning to transfer in-hand manipulations using a greedy shape curriculum. In: Computer Graphics Forum. vol.~42, pp. 25--36. Wiley Online Library (2023)

\bibitem{zhao2020sim}
Zhao, W., Queralta, J.P., Westerlund, T.: Sim-to-real transfer in deep reinforcement learning for robotics: a survey. In: 2020 IEEE symposium series on computational intelligence (SSCI). pp. 737--744. IEEE (2020)

\bibitem{zhuang2020comprehensive}
Zhuang, F., Qi, Z., Duan, K., Xi, D., Zhu, Y., Zhu, H., Xiong, H., He, Q.: A comprehensive survey on transfer learning. Proceedings of the IEEE  \textbf{109}(1),  43--76 (2020)

\end{thebibliography}

\clearpage

\appendix

\noindent\textbf{Overview.} 
The appendix contains a list of supplementary materials to support the main paper.

\begin{itemize}
    \item \textbf{Additional Technical Explanations (Section~\ref{sec:supp_method})}. We provide additional explanations to complement the main paper.
    \begin{itemize}
        \item \textit{Dexterous Manipulation Transfer (Section~\ref{sec:supp_method_task})}. We provide a more formal task definition, outlining its objectives and the involved functions.
        \item  \textit{Parameterized Quasi-Physical Simulators (Section~\ref{sec:supp_method_sim})}. Detailed explanations of the parameterized point set dynamics, including its full dynamic equations, and the parameterized residual physics, covering network designs, features, input, and output details.
        \item \textit{Dexterous Manipulation Transfer via a Physics Curriculum (Section~\ref{sec:supp_method_dex_transfer})}. We include comprehensive illustrations of the transfer process based on point sets, the iterative optimization procedure for approximating realistic dynamics, and detailed MPC procedure. 
    \end{itemize}
    \item  \textbf{Additional Experiments  (Section~\ref{sec:supp_exp})}. We present further experimental results to demonstrate the effectiveness of our method, along with discussions, analyses, additional comparisons, a user study and insights into failure cases and limitations.
    \begin{itemize}
        \item \textit{Discussions on Sim-to-Real and Real Robot Experiments (Section~\ref{sec:realrobotexp})}. Additional discussions on sim-to-real and real robot experiments. 
        \item \textit{Transferred Dexterous Manipulations (Section~\ref{sec:supp_additional_exp_results})}. Additional qualitative results showcasing intricate manipulations to highlight our method's capability.
        \item  \textit{Further Discussions and Analysis (Section~\ref{sec:supp_exp_discussions})}. We delve deeper into the role of MPC in our method, the question of does the residual physics module really compensates for the estimation other than taking the main role, the intermediate optimization processes in the quasi-physical simulator curriculum, and experiments on a different simulated robot hand whose morphology is significantly different from the human hand to demonstrate our method's capability in such cases. 
        \item \textit{Additional Comparisons (Section~\ref{sec:supp_additional_cmps})}. In addition to comparisons with previous Reinforcement Learning (RL) methods, we compare approaches that incorporate human demonstrations into policy learning for acquiring skills.
        \item \textit{Failure Cases (Section~\ref{sec:supp_failure})}. Analysis of failure cases to gain insights into limitations and areas for improvement.
        \item  \textit{User Study (Sec.~\ref{sec:supp_user_study})}. We additionally include a user study to further assess effectiveness of our method. 
    \end{itemize}
    \item \textbf{Experimental Details (Section~\ref{sec:supp_exp_details})}.  We illustrate details of datasets, metrics, baselines, models, evaluation settings, and running time as well as the complexity analysis. 
    \item  \textbf{Potential Negative Societal Impact (Section~\ref{sec:neg_social_impact}).} We discuss the potential negative social impacts of the work.  
\end{itemize}

We include a \textbf{\href{https://meowuu7.github.io/QuasiSim/}{website}} and a \textbf{\href{https://youtu.be/Pho3KisCsu4}{\textcolor{orange}{video}}} to introduce our work. The website and the video contain \textit{\textcolor{mypurpleee}{animated transferred dexterous manipulations}}. We highly recommend exploring these resources for an intuitive understanding of the task, difficulties, the effectiveness of our model, and its superiority over prior approaches.
We include source code in the supplemental material. We will publicly release the code and data upon acceptance of the paper.


\section{Additional Technical Explanations} \label{sec:supp_method}

\begin{figure*}[h]
  \centering
  \includegraphics[width=1.0\textwidth]{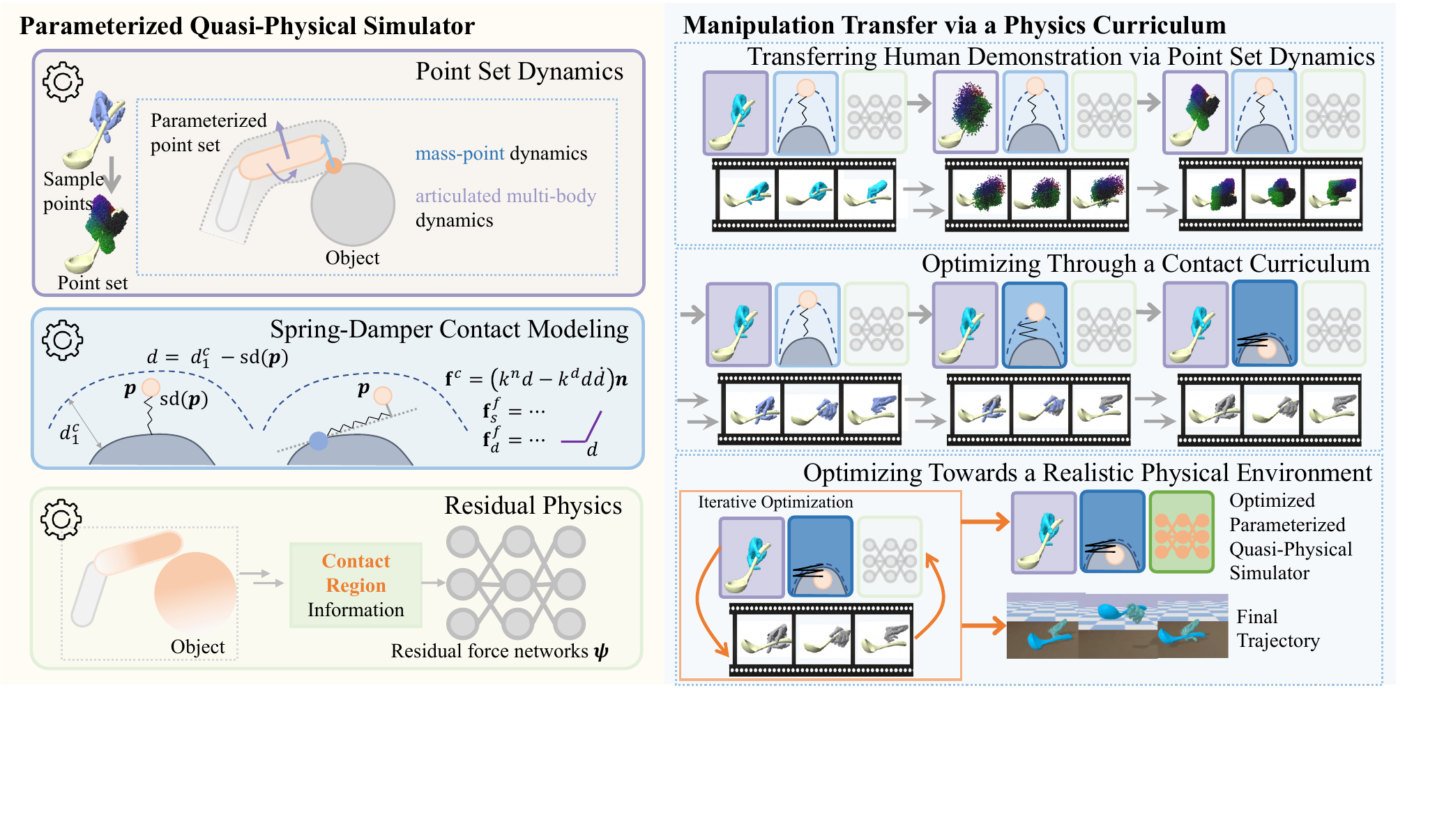}
  \vspace{-40pt}
  \vspace{-7pt}
  \caption{ 
  \textbf{Detailed Method Overview.}
  The \textbf{parameterized quasi-physical simulator} relaxes the articulated multi rigid body dynamics as the \textit{parameterized point set dynamics}, 
  controls the contact behavior via an unconstrained \textit{parameterized spring-damper contact model},
  and compensates for unmodeled effects via \textit{parameterized residual physics networks}. 
  We tackle the difficult dexterous manipulation transfer problem via \textbf{a physics curriculum}. 
  }
  \label{fig_supp_detailed_method_pipeline}
\end{figure*}

We include a figure providing a comprehensive overview of the method (see Fig.~\ref{fig_supp_detailed_method_pipeline}).

\subsection{Dexterous Manipulation Transfer} \label{sec:supp_method_task}

Given a human manipulation demonstration, composed of a human hand mesh trajectory and an object pose trajectory $\{ \mathcal{H} = \{ \mathbf{H}_n\}_{n=1}^{N}, \mathcal{O} = \{ \mathbf{O}_n \}_{n=1}^{N}  \}$ with $N$ frames, the goal is transferring the demonstration to a dexterous robot hand in simulation. Formally, we aim to optimize a control trajectory $\mathcal{A}$ that drives the dexterous hand to manipulate the object in a realistic simulated environment so that the resulting hand trajectory ${ \hat{\mathcal{H}} = \{  \hat{\mathbf{H}}_n \}_{n=1}^{N} }$ and the object trajectory  $\hat{\mathcal{O}} = \{ \hat{\mathbf{O}}_n \}_{n=1}^N$ are close to the reference motion $\{ \mathcal{H}, \mathcal{O} \}$. Since the object properties and the system parameters are unknown from the kinematics-only trajectory, we estimate such parameters, denoted as set $\mathcal{S}$, along with the hand control optimization. 

\begin{figure}[h]
  \centering
  \includegraphics[width=0.7\textwidth]{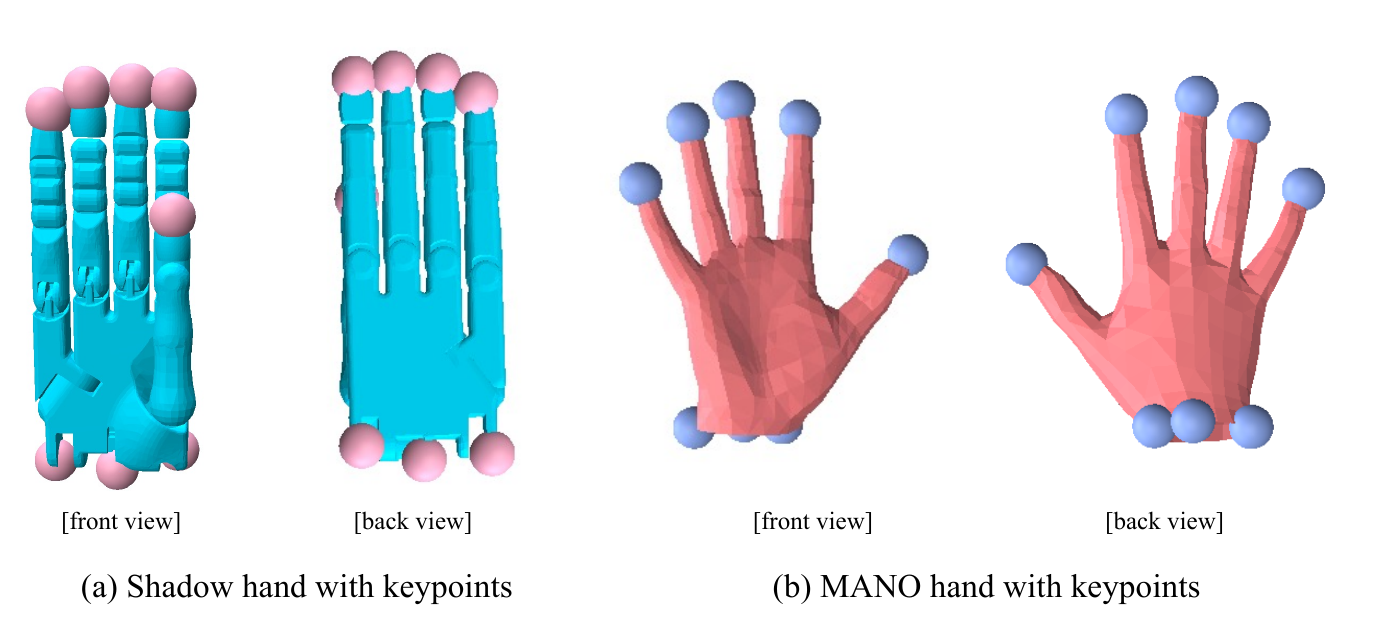}
  \caption{
  Hands with keypoints (keypoints are drawn as large \textcolor{pink}{pink} and \textcolor{mypurpleee}{blue purple} points).
  }
  \label{hand_with_keypoints}
\end{figure}

\noindent\textbf{Optimization objective.}
The task aims at optimizing a hand control trajectory  $\mathcal{A}$ so that the resulting hand trajectory $\hat{\mathcal{H}}$ and the object trajectory $\hat{\mathcal{O}}$ are close to the reference motions $\{ \mathcal{H} , \mathcal{O}\}$. Formally, the objective is:
\begin{equation}
    \text{minimize}_{\mathcal{A}, \mathcal{S}} w^o f^{\mathcal{O}}(\mathcal{O}, \hat{\mathcal{O}}) + w^h f^{\mathcal{H}}(\mathcal{H}, \hat{\mathcal{H}}),
\end{equation}
where $w^o$ and $w^h$ are object tracking weight and the hand tracking weight respectively, $f^{\mathcal{O}}$ measures the difference between two object pose trajectories, and $f^{\mathcal{H}}$ calculates the difference between two hand trajectory. 
Specifically, 
\begin{align}
    f^{\mathcal{O}} (\mathcal{O}, \hat{\mathcal{O}}) &= \frac{1}{N} \sum_{n=1}^{N}  ( (1 - \mathbf{q}_n\cdot \hat{\mathbf{q}}_n) + \Vert \mathbf{t}_n - \hat{\mathbf{t}}_n \Vert ) \\ 
    f^{\mathcal{H}} (\mathcal{H}, \hat{\mathcal{H}}) &= \frac{1}{N} \sum_{n=1}^{N} \Vert \mathbf{P}^{h}_n - \mathbf{P}^{r}_n \Vert ,
\end{align}
where $\mathbf{q}_n$ is the orientation of the $n$-th frame reference object pose, represented in quaternion, $\mathbf{t}_n\in \mathbb{R}^{3}$ is the translation of the $n$-th frame  reference object pose, $\hat{\mathbf{q}}_n$ and $\hat{\mathbf{t}}_n$ are the quaternion and the translation of the $n$-th frame estimated object pose, $\mathbf{P}_n^{h}$ is the reference human hand keypoint at the $n$-th frame, and $\mathbf{P}_n^{r}$ is the estimated robot hand keypoint at the $n$-th frame correspondingly. Keypoints consist of five fingertips and three points on the hand wrist. 
We manually defined them (Fig.~\ref{hand_with_keypoints}). 
Weights $w^o$ and $w^h$ are set to 1.0, 1.0 in our method. 



\subsection{Parameterized Quasi-Physical Simulators} \label{sec:supp_method_sim}

\begin{figure}[htbp]
  \centering
  \includegraphics[width=0.25\textwidth]{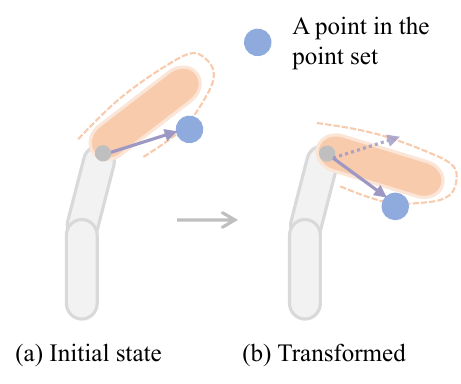}
  \caption{
   A point in the point set is regarded as ``attached'' to the body it sampled from and is affected by joint actions accordingly.
  }
  \label{fig_pointtranformations}
\end{figure}

\noindent\textbf{Parameterized point set dynamics.} 
Each point $\mathbf{p}_i$ in the point set $\mathcal{Q}$ is treated as a mass point with a finite mass $\mathbf{m}_i$ and infinitesimal volume. The action space of the point set is composed of the joint forces $\mathbf{u}\in \mathbb{R}^{n_r}$ in the reduced coordinate system, alongside a 3 degrees of freedom free force $\mathbf{a}_i\in \mathbb{R}^{3}$ applied to each point $\mathbf{p}_i \in \mathcal{Q}$. A point is considered to be ``attached'' to the body it was sampled from and can undergo articulated transformations, as illustrated in the example shown in Figure~\ref{fig_pointtranformations}.
The dynamics of the point set encompass articulated multi-body dynamics~\cite{Featherstone2007RigidBD,liu2012quick}, along with the mass point dynamics of each individual point $\mathbf{p}_i$.
Specifically, 
\begin{align}
    \mathbf{M}_r \ddot{\mathbf{q}}_r &= \tilde{\mathbf{f}}_r + (1 - \alpha) \mathbf{J}_{mr}^T \mathbf{f}_m + \mathbf{f}_{QVV} + \mathbf{u}, \\
    {m}_i \ddot{\mathbf{x}}_i &= \mathbf{J}_{i}\mathbf{u} +  \alpha \mathbf{f}_i + \alpha, \forall \mathbf{p}_i \in \mathcal{Q},
\end{align}
where $\mathbf{M}_r\in \mathbb{R}^{n_r\times n_r}$ is the generalized inertia matrix in reduced coordinates, $n_r$ is the number of freedom of the articulated object, $\mathbf{q}_r \in \mathbb{R}^{n_r}$ is the reduced state vector of the articulated object, $\tilde{\mathbf{f}}_r$ is the reduced force vector generated by joint-space such as joint damping and stiffness, $\mathbf{J}_{mr}$ is the Jacobian mapping generalized velocity $\dot{\mathbf{q}}_r$ to its maximal coordinate counterpart $\dot{\mathbf{q}}_m$, $\mathbf{f}_m$ is the maximal wrench vector including force and torque generated in maximal coordinate system, $\mathbf{f}_{QVV}$ is the quadratic velocity vector,  $\mathbf{u}$ denotes the generalized joint force, $\mathbf{J}_{i}$ represents the Jacobian mapping from the generalized velocity to the point velocity $\dot{\mathbf{x}}_i$, $\mathbf{f}_i$ accounts for external forces acting on $\mathbf{p}_i$, and $\mathbf{a}_i\in \mathbb{R}^{3}$ represents the actuation force applied to the point $\mathbf{p}_i$. Consequently, the point set is controlled by a shared control in the reduced coordinate space $\mathbf{u}$ and per-point actuation force $\mathbf{a}_i$. 

\noindent\textbf{Parameterized residual physics.} 
We introduce two residual contact force networks to compensate for the inherent limitations of the spring-damper based contact modeling. 
For detailed residual contact force prediction, we introduce a local contact network $f_{\psi_{\text{local}}}$ that utilizes contact information identified in the parameterized contact model and predicts residual forces between each contact pair. 
For each point pair in contact $(\mathbf{p}, \mathbf{p}^o)$, the local contact region is composed of $N_c^l$ object surface points and $N_c^l$ hand surface points. 
For the contact point in the object surface $\mathbf{p}^o$, we identify a region which contains object surface points whose distance to point $\mathbf{p}^o$ is not larger than a threshold $d_{thres}^l = 0.05$ (5cm) (point $\mathbf{p}^o$ is not included in the region). After that, $N_c^l - 1$ points are sampled from such points via farthest point sampling. These points, together with $\mathbf{p}^o$ are taken as the object local contact surface points. 
$N_c^l$ hand points are sampled in the same way. We set $N_c^l$ to 100 in experiments. 
After that, the local contact information consists of the geometry of the local contact region $\mathbf{P}_c^l\in \mathbb{R}^{2N_c^l\times 3}$, per-point velocity $\mathbf{V}_{c}^l\in \mathbb{R}^{2N_c^l\times 3}$, and per-object point normal $\mathbf{N}_c^l \in \mathbb{R}^{N_c^l\times 3}$. A PointNet is used to encode the contact region feature. The feature of each point is composed of the point type embedding vector (128 dimension), point position, point velocity, point normal (all zeros for hand points). The hidden dimensions are [128, 256, 512, 1024]. After that, we calculate the global feature via a `maxpool` operation. 
Then the global features is fed into the contact force prediction module for local residual contact force prediction. The prediction network is an MLP with hidden dimensions [512, 256, 128]. \texttt{ReLU} is leveraged as the activation layer.

To address discrepancies in contact region identification between the parameterized contact model and real contact region, we also incorporate a global residual network $f_{\psi_{\text{global}}}$ that predicts residual forces and torques applied directly to the object's center of mass.
To identify a global contact region, we adopt a similar way that first identifies a region on the object, containing object surface points whose distance to the nearest object contact point are smaller than the global contact distance threshold $d_{thres}^g = 0.1$ (10cm). After that, global contact region points are sampled for both the object and the hand in the same way as sampling the local contact region points described above. The number of global contact points on for the object and the hand is $N_c^g = 500$. Subsequently, the global contact region feature is encoded from the global contact region in the same way as does for local contact region feature. Then, the global contact feature is fed to a prediction network for predicting residual force and residual torque. The network architecture is the same as that for local residual force, but with a different output dimension (3 for force, 3 for torque, and 6 dimension in total).

\subsection{Dexterous Manipulation Transfer via a Physics Curriculum}  \label{sec:supp_method_dex_transfer}

\noindent\textbf{Transferring human demonstration via point set dynamics.}
The articulated rigid constraints are relaxed initially to facilitate robust manipulation transfer between two morphologically different robot hands and to overcome noise in the kinematic trajectory. After we have optimized the control trajectory of the point set constructed from the dynamic MANO hand~\cite{christen2022d}, the next goal is optimizing the control trajectory of the point set constructed from the simulated robot hand. Reliable correspondences between points are required to complete the transfer. Therefore, we first optimize the kinematics-only trajectory of the simulated robot hand based on coarse correspondences defined on keypoints (Fig.~\ref{hand_with_keypoints}). The objective is to track the MANO hand trajectory. After that, we define single directional point-point correspondence from the point set of the MANO hand to the point set of the simulated robot hand via the nearest neighbor. That is, for each point in the point set of the MANO hand, we find its nearest point in the point set of the simulated robot hand as its correspondence. After that, the hand tracking objective between the point set of the MANO hand and that of the simulated robot hand becomes the average distance between point-point in correspondence. Subsequently, the control trajectory of the point set is optimized so that the manipulated object pose trajectory can track the reference object pose trajectory, and the trajectory of the simulated robot hand's point set can track the trajectory of the MANO hand's point set. The control trajectory of the point set is first initialized via the kinematic trajectory of the point set via differentiable forward dynamics and optimization.


\noindent\textbf{Optimizing towards a realistic physical environment.}
When transferring to a realistic physical environment, we iteratively optimize the control trajectory $\mathcal{A}$ and the parameterized simulator. 
In more detail, in each iteration, the following steps are executed: 
\begin{itemize}
    \item Sample the replay buffer $\mathcal{B}$ from the interested realistic simulated environment. 
    \item Optimize the quasi-physical simulator to approximate realistic dynamics by ensuring that the simulated trajectory closely tracks the trajectory stored in the replay buffer.
    \item  Optimize the control trajectory $\mathcal{A}$ to accomplish the manipulation task within the quasi-physical simulator.
\end{itemize}

\noindent\textit{Tracking via closed-loop MPC.} 
After completing the optimization, 
the final control trajectory is yielded by model predictive control (MPC)~\cite{garcia1989model} based on the optimized parameterized simulator and the hand trajectory $\mathcal{A}$. Specifically, in each step $n$, the current $\mathbf{A}_n$ and the following controls in several subsequent frames $\{ \mathbf{A}_{n+1}, ..., \mathbf{A}_{n + q - 1} \}$ are optimized to reduce the tracking error. Denote the simulated object pose trajectory as  $\hat{\mathcal{O}}_n^{q} = \{ \hat{\mathbf{O}}_{n+1},...,\hat{\mathbf{O}}_{n+q} \}$, the corresponding reference object pose trajectory as $\mathcal{O}_n^{q} = \{ \mathbf{O}_{n+1}, ..., \mathbf{O}_{n+q} \}$, the simulated hand trajectory as $\hat{\mathcal{H}}_n^q = \{ \hat{\mathbf{H}}_{n+1}, ..., \hat{\mathbf{H}}_{n+q} \}$ with the corresponding keypoint trajectory $ \{ \mathbf{P}^{r}_{n+1}, ..., {\mathbf{P}}^{r}_{n+1}\}$ and reference hand keypoint trajectory $\{ \mathbf{P}^{h}_{n+1}, ..., {\mathbf{P}}^{h}_{n+1}\}$ the objective at each step $n$ is as follows:
\begin{equation}
    \text{minimize}_{\mathcal{A}} w^o f^{\mathcal{O}} (\hat{\mathcal{O}}_n^{q}, \mathcal{O}_n^{q}) + w^h f^{\mathcal{H}} ( \{ \mathbf{P}^{r}_{n+1}, ..., {\mathbf{P}}^{r}_{n+1} \}, \{ \mathbf{P}^{h}_{n+1}, ..., {\mathbf{P}}^{h}_{n+1} \} ).
\end{equation}
We update the control trajectory to minimize the objective via 10 steps gradient descent with a learning rate $10^{-4}$. 



\section{Additional Experiments} \label{sec:supp_exp}







In this section, we present additional experimental results that delve into more qualitative results on challenging cases (see  Section~\ref{sec:supp_additional_exp_results}), further analysis and discussions (see~ Section  \ref{sec:supp_exp_discussions}), additional comparisons (see Section~\ref{sec:supp_additional_cmps}), failure case analysis (see  Section~\ref{sec:supp_failure}), and a user study (see Section~\ref{sec:supp_user_study}).
Initially, we present additional experimental results achieved by our approach to further demonstrate its effectiveness. 
Subsequently, we delve into further discussions, including the role of MPC in our method, further investigations in the residual physics module, the intermediate optimization processes in the quasi-physical simulator curriculum, and experiments conducted on a different simulated robot hand that suffers from a significant morphology difference from the human hand. 
Then we present additional comparisons to the literature where human demonstrations are incorporated into policy learning. 
After that, we discuss failure cases and analyze our limitations. 
At last, we present a toy user study as an additional evaluation. 


\subsection{Discussions on Sim-to-Real and Real Robot Experiments} \label{sec:realrobotexp}

\begin{figure}[h]
\begin{center}
\vspace{-3pt}
   \includegraphics[width=\linewidth]{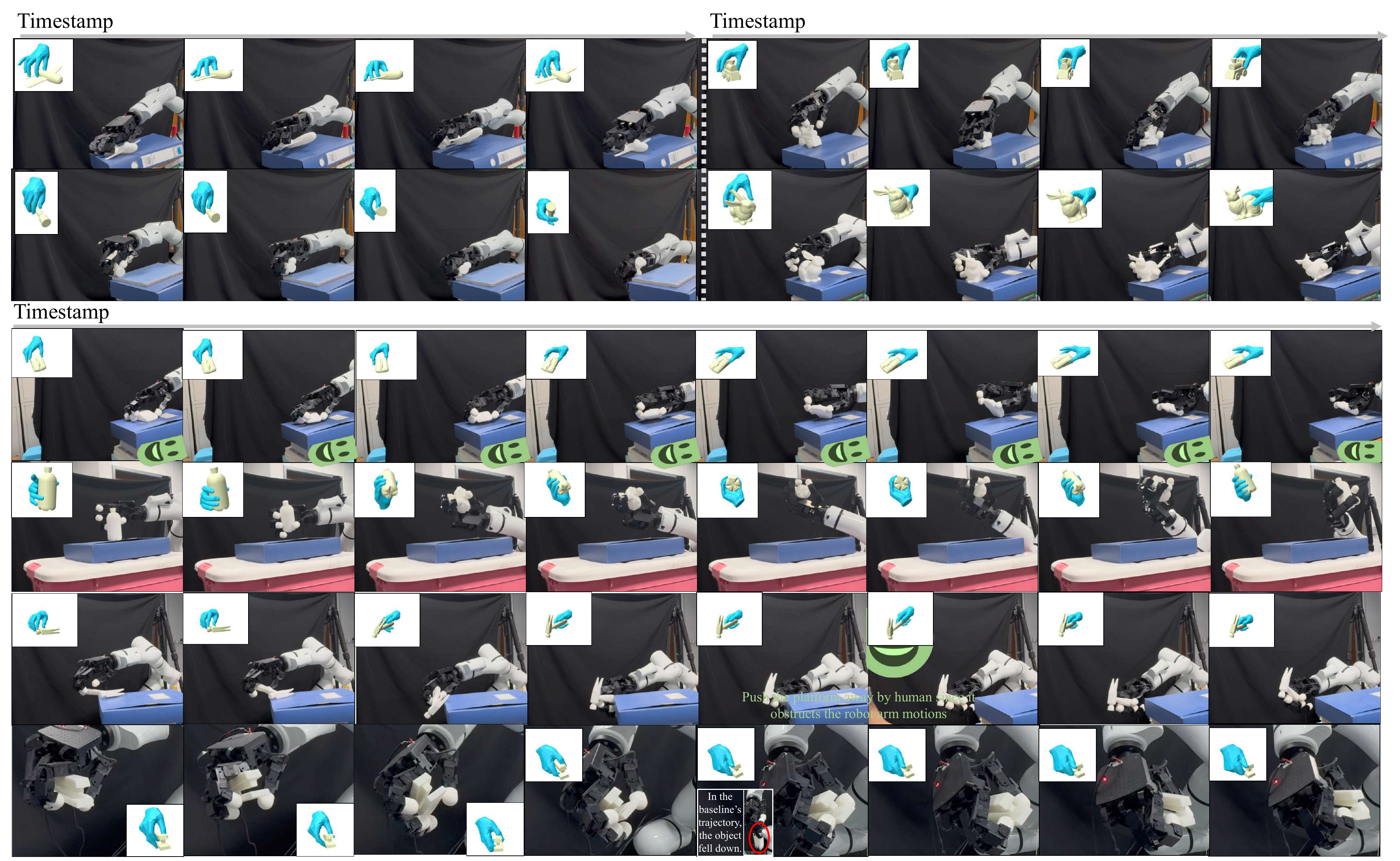}
\end{center}
\vspace{-20pt}
   \caption{\footnotesize Qualitative results on a real Allegro hand.  Please visit \textbf{\href{https://meowuu7.github.io/QuasiSim/}{our website}} for animated demonstrations. 
   }
\vspace{-20pt}
\label{fig:real_exp}
\end{figure}


\noindent\textbf{Sim-to-real challenges and possible solutions.}
Sim-to-real gaps primarily stem from differences in physics and system parameters between simulators and the real world. A straightforward strategy is ``direct sim-to-real'', \emph{i.e.,} optimizing in a realistic simulator and directly transferring results to a real robot. A more promising way is to train the quasi-physical simulator using real robot trajectories, acquired using offline policies, to approximate the real physics, followed by planning within it. Another approach is iterative quasi-physical simulator optimization and real robot executions. It can possibly learn real physics better but is expensive and faces safety issues. 

\noindent\textbf{Real robot experiments.}
Due to the high cost of Shadow hand hardware, we conducted experiments on a real Allegro hand and 3D printed objects to demonstrate real robot effectiveness. 
We adopt the ``direct sim-to-real'' for its simplicity, which transfers Allegro trajectories optimized in PyBullet to the real robot. We compared our method with DGrasp-Tracking on 12 well-tracked trajectories in the simulator, some with complex contacts and large object movements. Our method succeeded in 8 out of 12 trajectories without dropping the object (Fig.~\ref{fig:real_exp}), while the baseline only succeeded in 4. This suggests the potential value of our method for real robot applications. Using advanced strategies proposed above may further improve the performance. 



\subsection{Transferred Dexterous Manipulations} \label{sec:supp_additional_exp_results}

\begin{figure}[htbp]
  \centering
  \includegraphics[width=0.8\textwidth]{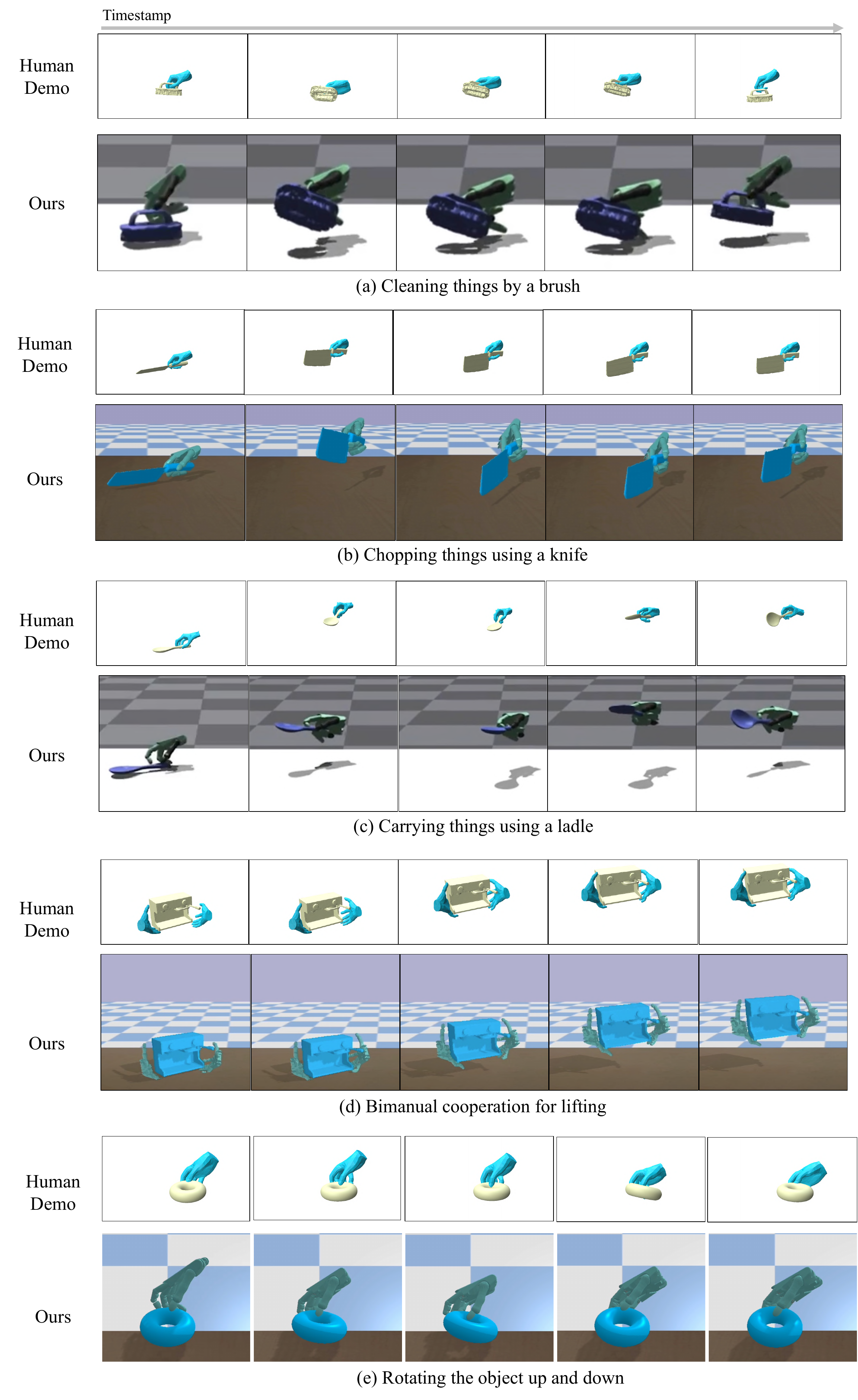}
  \caption{
  \textbf{Transferred manipulations.} We provide additional examples to demonstrate the effectiveness of our method. Our approach successfully tracks complex manipulations involving subtle object movements, such as gently shaking a brush for cleaning (\textit{Fig. (a)}), employing non-trivial functional tools (\textit{Fig. (b) (c) (e)}), and executing bimanual cooperation tasks (\textit{Fig. (d)}). For animated demonstrations, please visit \textbf{\href{https://meowuu7.github.io/QuasiSim/}{our website}} and refer to the \href{https://youtu.be/Pho3KisCsu4}{\textcolor{orange}{accompanying video}}.
  }
  \label{fig_supp_more_exp_res}
\end{figure}


Figure~\ref{fig_supp_more_exp_res} showcases supplementary experimental results obtained through our method. We highly encourage readers to explore our \textbf{\href{https://meowuu7.github.io/QuasiSim/}{website}} and view the \href{https://youtu.be/Pho3KisCsu4}{\textcolor{orange}{accompanying supplementary video}} for animated demonstrations.


\subsection{Further Discussions and Analysis} \label{sec:supp_exp_discussions}

\begin{figure}[htbp]
  \centering
  \includegraphics[width=\textwidth]{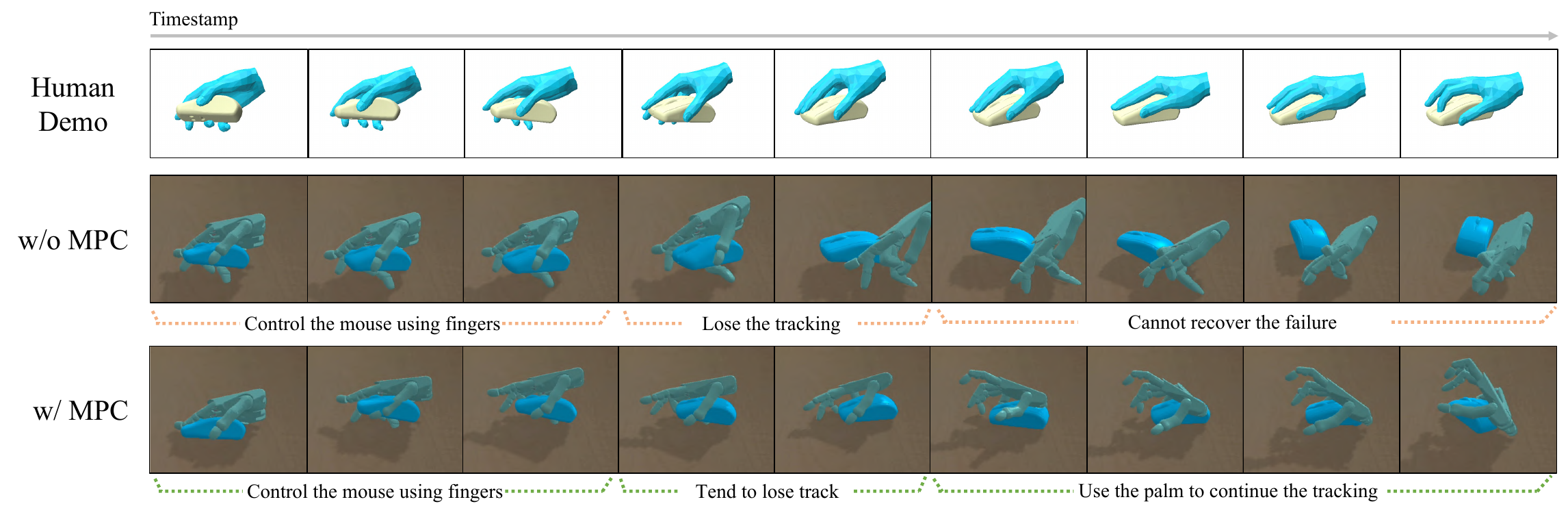}
  \caption{
  \textbf{Robustness of MPC. } MPC tries to track the object even after experiencing a dangerous period with the tendency to lose track. While the trajectory yielded by open-loop optimization fails. 
  }
  \label{fig_mpc_robustness}
\end{figure}

\noindent\textbf{Robustness of MPC.} 
Fig.~\ref{fig_mpc_robustness} shows an example demonstrating tracking robustness. In this challenging example where rich contacts between fingers and the palm with the mouse are frequently established and broken, the control sequence optimized in an open-loop manner struggles with keep contacting the mouse, and the tracking is lost finally. However, with the optimized model, the trajectories produced by MPC can successfully maintain enough contact with the object and track the sequence naturally.



\begin{figure}[h]
  \centering
  \includegraphics[width=0.8\textwidth]{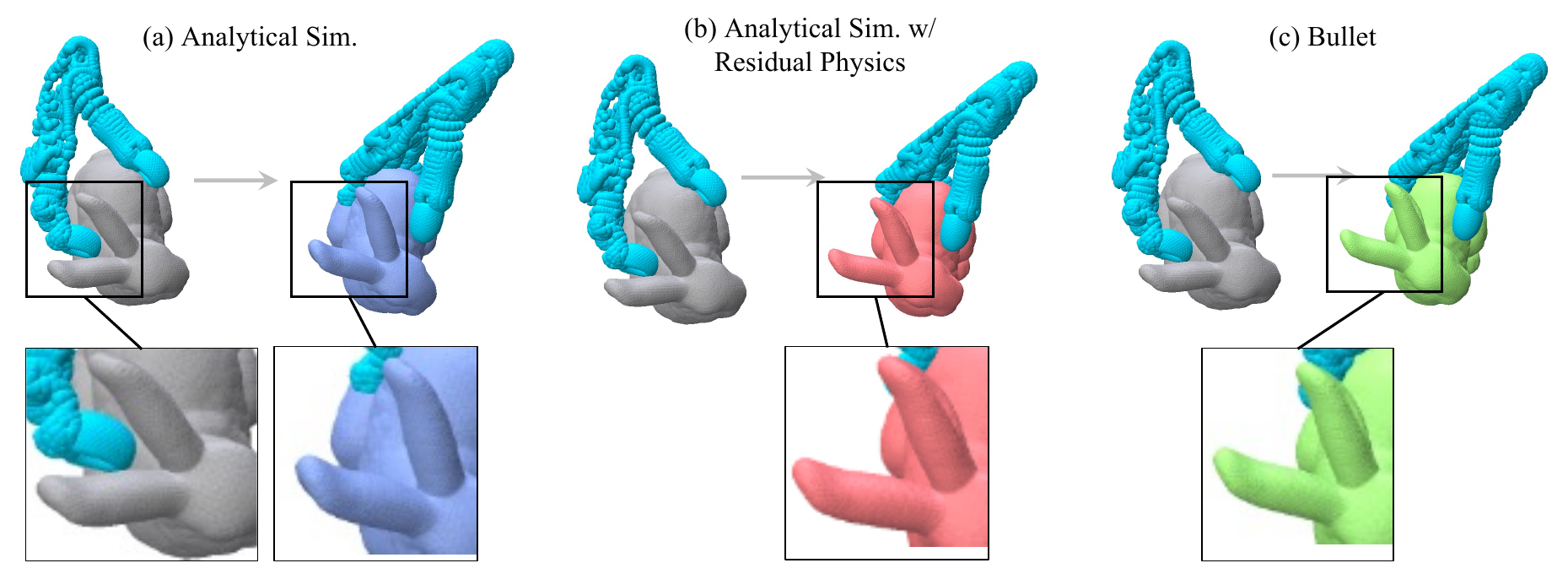}
  \caption{
  \textbf{Analysis on the residual physics module.} In this 10-step transition, the transformed bunny predicted by the analytical part of the quasi-physical simulator (\textcolor{mypurpleee}{purple bunny}) only is already close to the GT one (\textcolor{green}{green} bunny). The residual physics can compensate for some unmodeled effects. Hence the result (\textcolor{red}{red bunny}) yielded by the quasi-physical simulator with both the analytical part and the residual physics module gets closer to the observation in Bullet. 
  }
  \label{fig_res_phy_bunny_trans}
\end{figure}

\begin{figure}[h]
  \centering
  \includegraphics[width=0.5\textwidth]{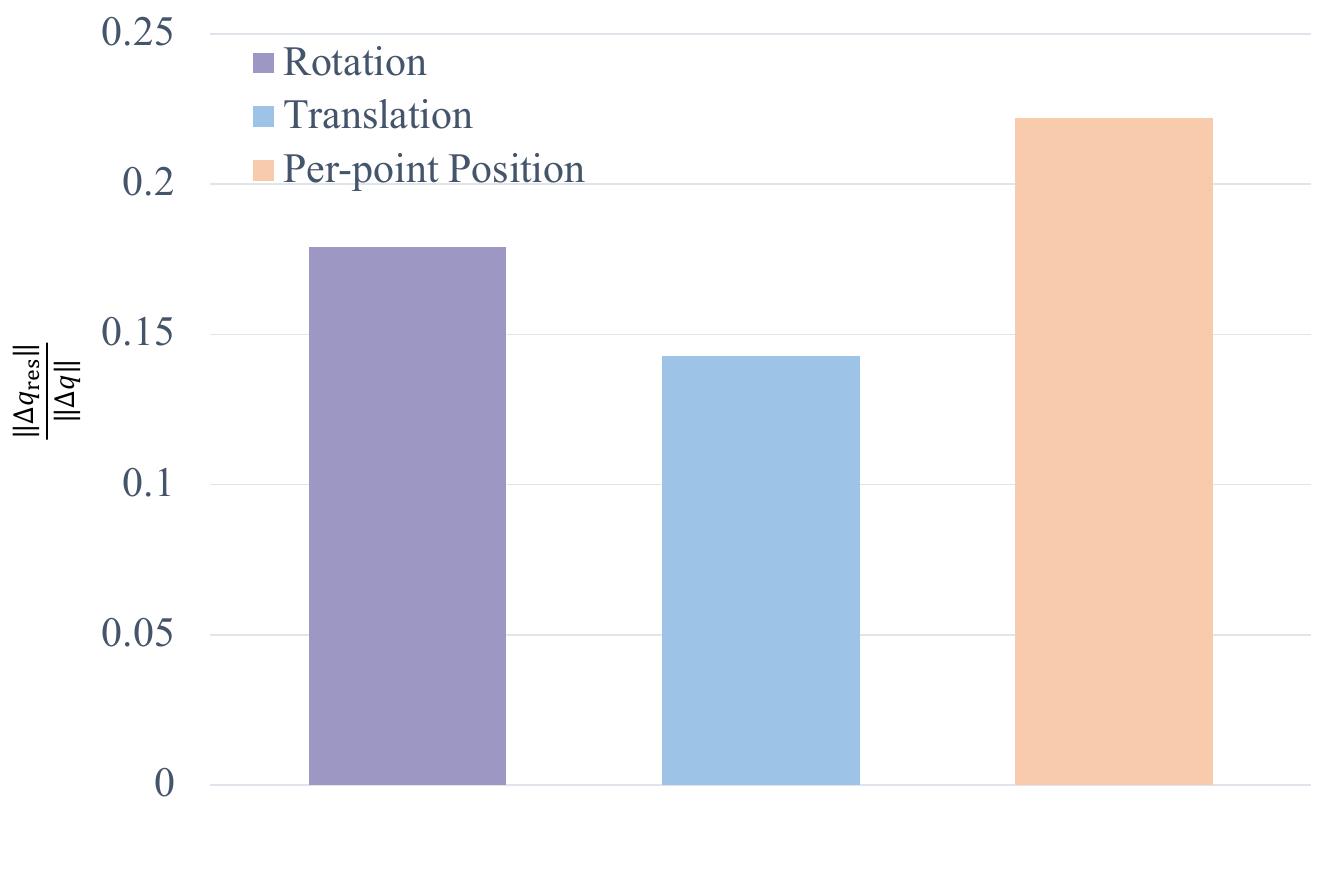}
  \caption{
  Quantitative analysis on the residual physics module.
  }
  \label{fig_res_phy_bunny_trans_quant}
\end{figure}

\noindent\textbf{Role of residual physics in quasi-physical simulators.} 
We evaluate the role of the residual physics on a small subset of our data from the GRAB dataset. 
We assess the impact of residual physics on a limited subset of our data from the GRAB dataset. This subset comprises 60 ten-step transitions involving manipulation sequences with objects such as \texttt{bunny}, \texttt{mouse}, \texttt{stapler}, \texttt{pyramid}, \texttt{cylinder}, \texttt{flashlight}, \texttt{watch}, \texttt{waterbottle}, \texttt{hammer}, and \texttt{clockarlam}.

To investigate whether the residual physics compensates for prediction while the analytical simulation remains predominant, we utilize two types of models: one comprising only the analytical part, and the other incorporating both the analytical part and the residual physics network. These models are tasked with predicting the object rotation and translation for each ten-step transition based on the object's initial state and hand action sequence.
Let $\mathbf{R}$ denote the object rotation predicted by the analytical part, and $\mathbf{R}_{\text{tot}}$ represent the rotation predicted by the analytical part with the residual model. Similarly, let $\textbf{t}$ and $\mathbf{t}_{\text{tot}}$ denote the object translation predicted by the analytical part and the analytical part with the residual model, respectively. 
Therefore, the residual rotation is calculated as $\mathbf{R}_{\text{res}} = \mathbf{R}_{\text{tot}} \mathbf{R}^T$, and the residual translation is calculated as $\mathbf{t}_{\text{res}} = \mathbf{t}_{\text{tot}} - \mathbf{R}_{\text{res}} \mathbf{t}$.
Let $\mathbf{V}_{\text{init}}$ denote the initial object vertices, $\mathbf{V}$ represent the transformed vertices, and $\mathbf{V}_{\text{tot}}$ denote the transformed vertices predicted by the analytical part with the residual model.
The average per-vertex position difference from the transformed object to the initial object is calculated as
\begin{equation}
    p_{\text{diff}} = \frac{1}{N_{v}} \Vert \mathbf{V} - \mathbf{V}_{\text{init}} \Vert.
\end{equation}
Similarly, the average per-vertex position difference from the transformed object predicted by the total model to the initial object is computed as
\begin{equation}
    p_{\text{diff}}^{\text{tot}} = \frac{1}{N_{v}} \Vert \mathbf{V}_{\text{tot}} - \mathbf{V}_{\text{init}} \Vert.
\end{equation}
Finally, the average per-vertex position difference from the transformed object predicted by the total model to the object predicted by the analytical part is calculated as:
\begin{equation}
     p_{\text{diff}}^{\text{res}} = \frac{1}{N_v} \Vert \mathbf{V}_{\text{tot}} - \mathbf{V} \Vert.
\end{equation}
For each 10-step transition, we calculate the relative quantities of the three types predicted by the residual physics, including the object rotation (measured by angles) $\frac{\text{angle}(\mathbf{R}_{\text{res}})}{\text{angle}(\mathbf{R}_{\text{tot}})}$, object translation $\frac{\mathbf{t}_{\text{res}}}{\mathbf{t}_{\text{tot}}}$, and the object per-point difference $\frac{p_{\text{diff}}^{\text{res}}}{ p_{\text{diff}} }$, compared to the overall predicted values by the quasi-physical simulator. 

As depicted in the bar chart shown in Figure~\ref{fig_res_phy_bunny_trans}, it is evident that the analytical model plays the primary role in predicting state transitions, while the information predicted by the residual module compensates for the prediction.
A visual example is depicted in Figure~\ref{fig_res_phy_bunny_trans}. The \texttt{bunny} undergoes rotation by a certain angle in the 10-step transition. The predicted result by the analytical part only is close to the ground-truth transformed object already. This alignment can be readily observed by examining the angle between the two ears of the bunny and the vertical/horizontal line, respectively. The residual physics compensate for unmodeled effects. Hence the object predicted by the full model is closer to the ground-truth transition observed in Bullet.




\begin{figure}[htbp]
  \centering
  \includegraphics[width=0.8\textwidth]{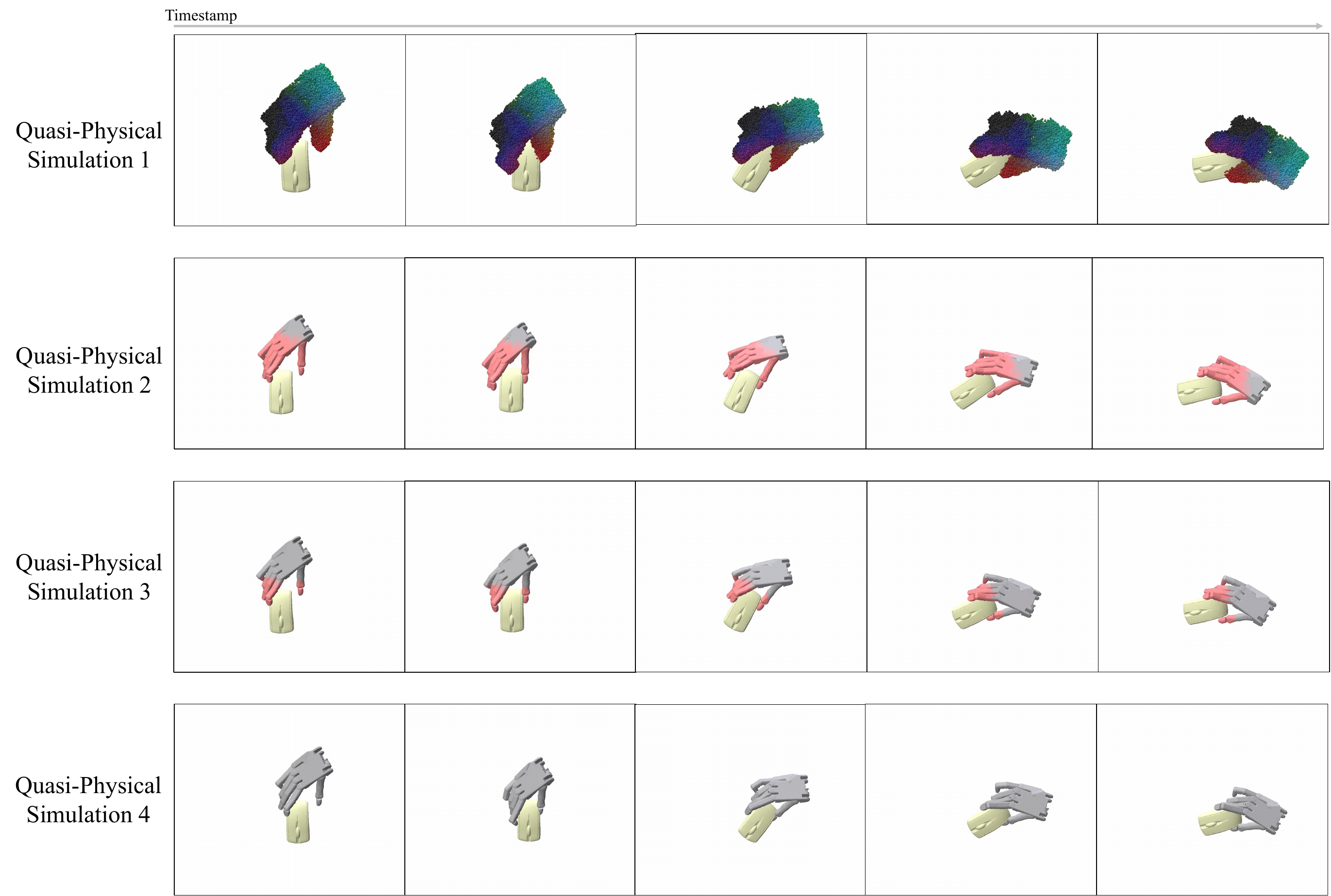}
  \caption{
  {Example of the optimization process in the \textbf{quasi-physical simulator curriculum}.}
Initially, both the contact constraints and the articulated rigid constraints are relaxed and the object is represented as a point set (the first line). Then the articulated rigid constraints are imposed and the contact model is gradually tightened. The optimization is solved in each of the simulators in the curriculum. We use \textcolor{myredddd}{orange red} color to represent the ``activated manipulators''.
  }
  \label{fig_quasi_physical_opt_process}
\end{figure}

\noindent\textbf{The optimization process in the quasi-physical simulator curriculum.} 
The quasi-physical simulator curriculum initially relaxes various constraints within the simulator to alleviate the optimization problem. Subsequently, the physics constraints are gradually tightened to enable the optimization to converge towards a solution in a more realistic physics model. Fig.~\ref{fig_quasi_physical_opt_process} illustrates the intermediate optimization process. 

During the first optimization iteration, articulated rigid constraints are relaxed, and the articulated rigid dexterous hand is represented and driven as a point set. Then, articulated constraints are imposed. The optimization continues in the simulator with an increasing contact stiffness (the following three lines in Fig.~\ref{fig_quasi_physical_opt_process}). 

Since the articulated dexterous hand is initially represented as a point set, comprised of points sampled from the ambient space of the surface mesh, contact between the hand and the object may not necessarily be established immediately. This is because contact can occur between points that are distant from each other, and these points can still act as manipulators. However, even with the articulated constraints removed during the initial optimization stages, the optimization process can still be effectively solved due to the softness of the contact model at the beginning.

As the optimization progresses, we gradually transition towards the final quasi-physical simulator with articulated rigid constraints and the stiffest contact model. In Fig.~\ref{fig_quasi_physical_opt_process}, we use \textcolor{myredddd}{orange red} color to represent the ``activated manipulators'' — surface points where contact can be established between them and the object.


\begin{figure}[h]
  \centering
  \includegraphics[width=0.9\textwidth]{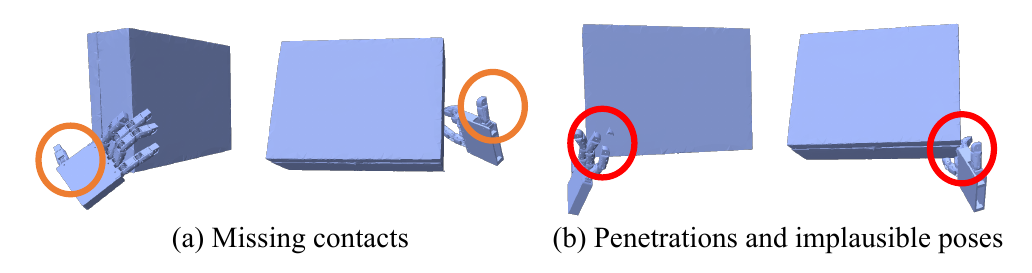}
  \caption{
  \textbf{Functionally implausible} transferred poses via sparse correspondences defined by keypoints. 
  }
  \label{fig_morphologicaly_different_robo_hand}
\end{figure}

\begin{figure}[h]
  \centering
  \includegraphics[width=\textwidth]{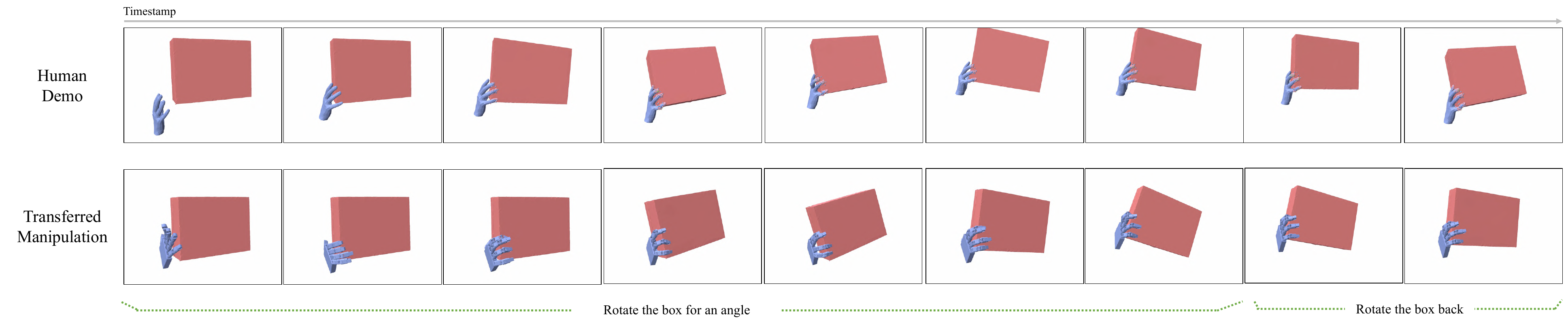}
  \caption{
  Manipulations transferred to a morphologically different dexterous robot hand. Taking advantage of the point set representation, the manipulation can be easily transferred to a dexterous hand with an extremely short thumb, which is different from the original MANO hand. 
  }
  \label{fig_morphologicaly_different_robo_hand_transfer_res}
\end{figure}

\begin{figure}[h]
  \centering
  \includegraphics[width=0.5\textwidth]{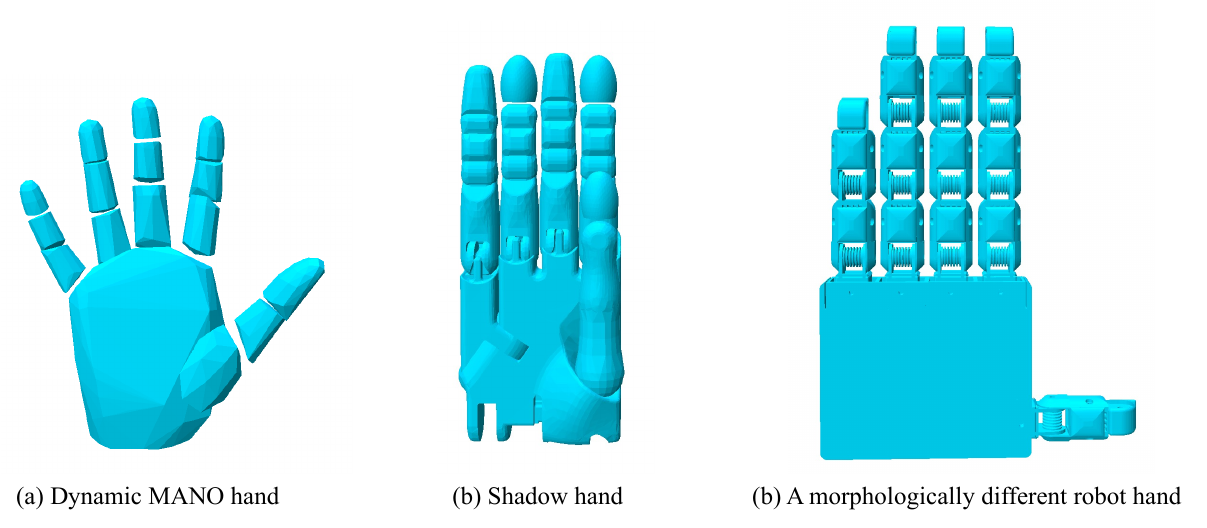}
  \caption{
  Comparisons between the dynamic MANO hand (\textit{Fig. (a)}) and two simulated robot hands (\textit{Fig. (b) (c)}) we considered in this work. Compared to the hand shown in Fig. (c), the Shadow hand is more similar to the human hand, but still with morphology differences that cannot be ignored. For fine-grained manipulation tasks, such morphological difference poses significant challenges for transferring. The hand in Fig. (c) is featured by its extremely short thumb and four other fingers longer than the human hand. Transferring human demonstrations to this hand is therefore very difficult. Our flexible point set representation, however, can still work in this case. 
  }
  \label{fig_hand_shapes}
\end{figure}


\noindent\textbf{Transferred to a robot hand with a significant morphological difference from the human hand.} 
Utilizing the point set representation, we can facilitate the transfer of manipulation skills to a morphologically different hand. We conducted additional experiments aimed at transferring manipulation from a human hand to a morphologically different robot hand obtained from DiffHand~\cite{xu2021end}. 
As shown in Fig.~\ref{fig_hand_shapes} (c), the thumb of the dexterous hand is obviously shorter than the human hand. Intuitively, completing manipulations using this hand is difficult. Directly transferring the manipulation via sparse correspondences defined between such two hands (\emph{e.g.,} finger and wrist correspondences as we have defined between the Shadow hand and the human hand (Fig.~\ref{hand_with_keypoints})) is not sufficient, leading to missing contacts and unwanted penetrations shown in Fig.~\ref{fig_morphologicaly_different_robo_hand}. However, as shown in Fig.~\ref{fig_morphologicaly_different_robo_hand_transfer_res}, our method can still effectively control it to complete the box rotation manipulation. Experiments are conducted in the last quasi-physical simulator from the curriculum.

\subsection{Additional Comparisons} \label{sec:supp_additional_cmps}

\begin{table*}[t]
    \centering
    \caption{ 
    \textbf{Additional Comparisons.}   
    Quantitative comparisons between our method and DexMV. Experiments are conducted on sequences from the GRAB dataset in the Bullet simulator.
    \bred{Bold red} numbers for best values.
    } 
\begin{tabular}{@{\;}lccccc@{\;}}
        \toprule

        ~ & \multicolumn{2}{c}{Object} & \multicolumn{2}{c}{Hand} &  \multicolumn{1}{c}{Overall}    \\
        \cmidrule(l{2pt}r{2pt}){2-3}
        \cmidrule(l{2pt}r{2pt}){4-5}
        \cmidrule(l{2pt}r{2pt}){6-6}
        
        Method & \makecell[c]{$R_{\text{err}}$ ($^\circ, \downarrow$)} & \makecell[c]{$T_{\text{err}}$ (${cm}, \downarrow$)}  &  MPJPE (${mm}, \downarrow$)  & CD (${mm}, \downarrow$) & Success Rate ($\%, \uparrow$)    \\

        \cmidrule(l{0pt}r{1pt}){1-1}
        \cmidrule(l{2pt}r{2pt}){2-6}

        \makecell[l]{DexMV} & 28.36 & 2.42 & 41.53 & 18.09 &  11.11/18.52/48.15
        \\ 
        
        Ours & \bred{22.38} & \bred{1.76} & \bred{35.02} & \bred{13.62} & \bred{25.93}/\bred{37.04}/\bred{62.96}
        \\ 
        \bottomrule
 
    \end{tabular}
    \label{tb_supp_additional_cmps}
\end{table*}

\begin{figure}[h]
  \centering
  \includegraphics[width=\textwidth]{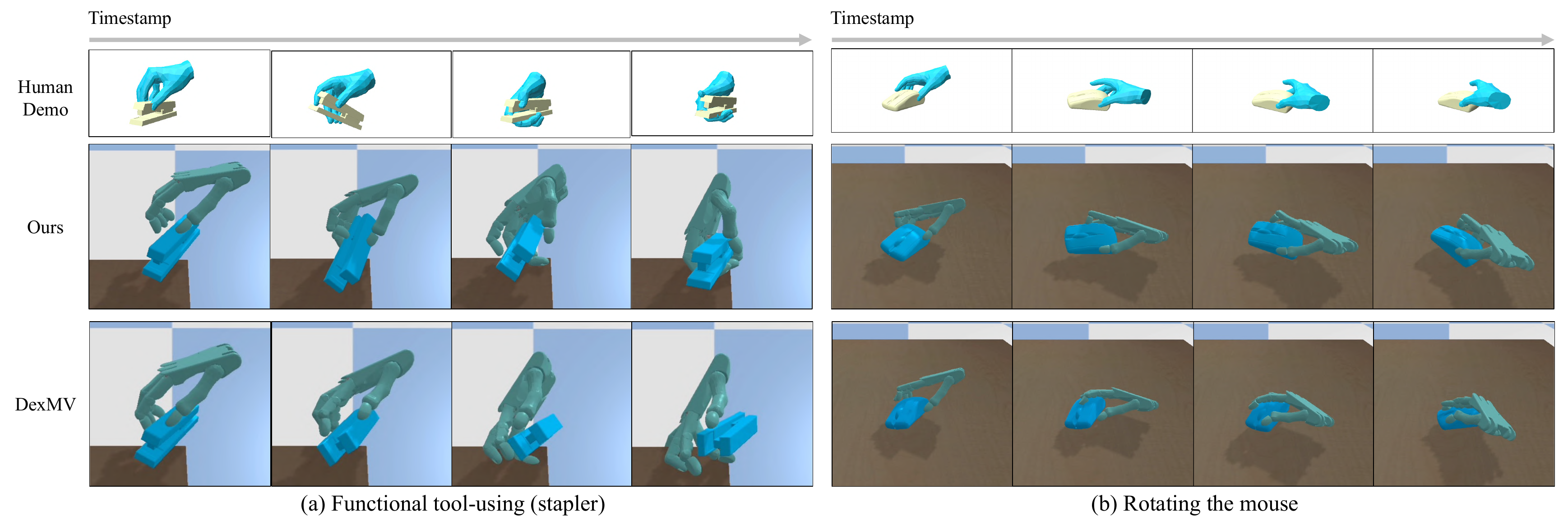}
  \caption{
  \textbf{Visual comparisons} between our method and DexMV. We can complete the tracking in a human-like way. However, DexMV cannot fulfill this vision. Its resulting trajectory may deviate from the human demonstration obviously, as observed in both Fig. (a) and (b). Besides, it struggles with the challenging example shown in Fig. (a) with rich and changing contacts. 
  }
  \label{fig_cmp_ours_dexmv}
\end{figure}

The main paper includes comparisons with both model-based and model-free approaches for solving the manipulation transfer task. For model-free methods, we compare with the DGrasp series models. The DGrasp series employs a carefully designed RL-based method for grasping, incorporating well-devised rewards containing position-to-goal information and contact information. Notably, DGrasp's methodology serves as the foundation for their recent work, ArtiGrasp~\cite{zhang2023artigrasp}. ArtiGrasp extends the manipulation capabilities to articulated objects and introduces learning techniques such as a gravity curriculum to handle complex relocate-and-articulate task settings. Given the meticulous reward design, stage-wise learning approach, and subsequent improvements, we consider DGrasp as a robust RL-based baseline. However, DGrasp is not explicitly designed for the tracking task, as it relies solely on sparse reference frames obtained from human demonstrations.  Therefore, we introduce the improved version of DGrasp-Tracking as our baseline. 


Many works have explored the combination of RL and imitation learning to leverage human demonstrations for learning robotic manipulation skills~\cite{qin2022dexmv,bahl2023affordances,bharadhwaj2023towards,chen2023human,guo2023learning,qin2023dexpoint,shaw2023videodex}. In these approaches, human demonstrations are utilized either as dense information for the robot to imitate or as sparse reward signals, such as grasp affordances~\cite{bahl2023affordances}. However, these methods often struggle with the imbalance between human-likeness and task completion, leading to biases towards RL-preferred trajectories.

For the sake of experimental completeness and to showcase the effectiveness of our strategy in contrast to this trend, we compare our approach with DexMV~\cite{qin2022dexmv}. Among its follow-ups and related works~\cite{qin2023dexpoint,arunachalam2023dexterous,bahl2023affordances}, DexMV shares the most similar setting to ours. In DexMV, human demonstrations provide dense references to shape the reward space for their RL algorithm. Furthermore, DexMV is openly available, making it conducive for comparative evaluation\footnote{\href{https://github.com/yzqin/dexmv-sim?tab=readme-ov-file}{DexMV's GitHub Repository Link}}.

We compare our method with DexMV (DAPG) on a subset, containing manipulation sequences from the GRAB dataset, in the Bullet simulator. Table~\ref{tb_supp_additional_cmps} presents the average quantitative results over the tested sequences. Fig.~\ref{fig_cmp_ours_dexmv} further leverages some examples to give an intuitive evaluation. 
In the challenging example shown in Fig.~\ref{fig_cmp_ours_dexmv} (a) with rich and changing contacts, our method can perform well. However, DexMv struggles to give satisfactory results. In the example shown in Fig.~\ref{fig_cmp_ours_dexmv} (b), we can track the object in a human-like way. However, though DexMV can complete the object tracking task to some extent, the resulting hand trajectory significantly deviates from the human hand demonstration. 





\begin{figure}[htbp]
  \centering
  \includegraphics[width=0.7\textwidth]{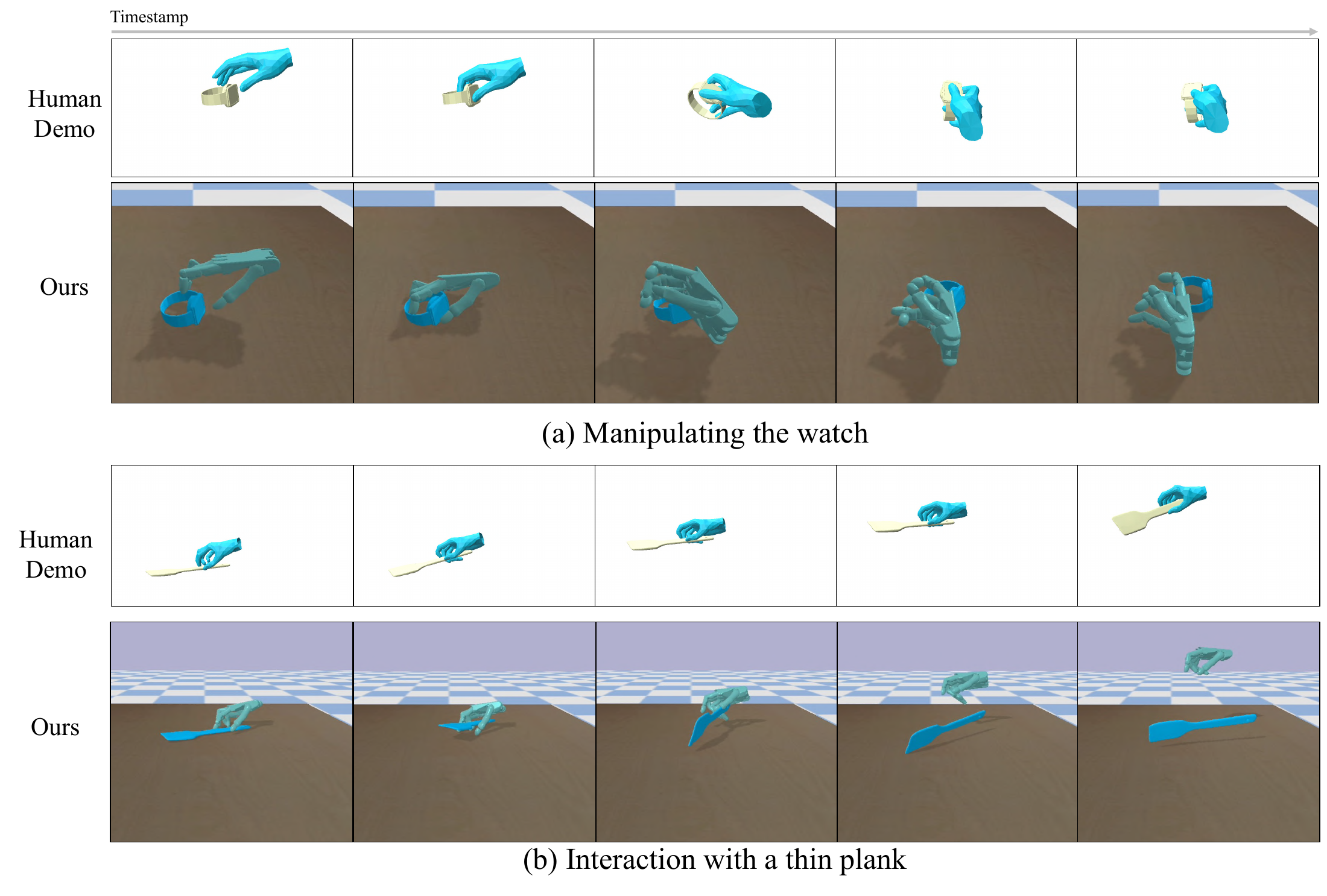}
  \caption{
  \textbf{Failure cases analysis.} \textit{Fig. (a)}: The hand fails to grasp the wristwatch, which requires us to control several fingers to pass through the ring of the wristwatch. \textit{Fig. (b)}: The hand fails to find a good strategy for lifting the thin plank. 
  }
  \label{fig_failurecases_watch}
\end{figure}

\subsection{Failure Cases} \label{sec:supp_failure}

In this section, we delve into the failure cases encountered by our method despite its effectiveness on many sequences. Our method may falter in controlling a simulated robot hand to track manipulation demonstrations in the following scenarios:
\begin{itemize}
    \item Manipulations requiring highly precise control, such as threading fingers through a ring for future actions (Fig.~\ref{fig_failurecases_watch} (a));
    \item Interactions with a nearly two-dimensional, very thin object (Fig.~\ref{fig_failurecases_watch} (b)). 
\end{itemize}
As depicted in Fig.~\ref{fig_failurecases_watch} (a), effectively controlling multiple fingers of the hand to pass through the ring of a wristwatch for secure attachment to the palm poses a significant challenge. Presently, our method struggles to provide satisfactory solutions for such cases, possibly due to morphological disparities between the human hand and the robot hand. These differences make it difficult to replicate human-like actions with the robot hand.
Additionally, we encounter difficulties achieving desirable outcomes when interacting with extremely thin objects, especially when one dimension of the object scales down to near-zero, as illustrated in Fig.~\ref{fig_failurecases_watch} (b). Such challenging object shapes make it challenging to devise an effective lifting strategy.

\subsection{User Study} \label{sec:supp_user_study}

We conduct a supplementary user study to complement the quantitative and qualitative evaluations presented in the main paper, website, and supplementary video, aiming to comprehensively assess and compare the quality of our transferred manipulations with those of the baseline method, DGrasp-Tracking.
Our user study is hosted on a website, where the results of our method and DGrasp-Tracking on 10 sequences are presented in a randomly permuted order. Ten participants, regardless of their familiarity with the task or expertise in computer science, are asked to rate each clip on a scale from 1 to 5 to indicate their preferences.
Specifically, ``1'' indicates a significant difference between the transferred motion and the reference motion, ``3'' represents the manipulation task is completed to some extent but the hand motion deviates obviously from the reference motion, ``5'' indicates a delicately controlled motion with a good task completeness and human-likeness. Intermediate values of "2" and "4" represent in-between assessments.


For each clip, we calculate the average score achieved by our method and DGrasp-Tracking. The average and median scores across all clips are summarized in Table~\ref{tb_exp_user_study}. The results show the significant superiority of our method over the baseline method.
\begin{table*}[t]
    \centering
    \caption{ 
    \textbf{User study.} 
    } 
\begin{tabular}{@{\;}lcc@{\;}}
        \toprule
        ~ & ~~~Ours~~~ & \makecell[c]{DGrasp- \\ Tracking  }
        \\
        \midrule
        \multirow{1}{*}{Average Score} & \textbf{4.00} & 2.06
        \\ 
        \midrule
        \multirow{1}{*}{Median Score} & \textbf{3.95} & 2.10
        \\ 
        \bottomrule
    \end{tabular}
    \label{tb_exp_user_study}
\end{table*} 

\section{Experimental Details} \label{sec:supp_exp_details}

\subsection{Datasets}

Evaluation data comes from three datasets, namely GRAB~\cite{taheri2020grab}, containing single-hand interactions with daily objects, TACO~\cite{liu2024taco}, containing humans manipulating tools, and ARCTIC~\cite{fan2023arctic} with bimanual manipulations. 
We'll publicly release the dataset for future research. 

\noindent\textbf{GRAB~\cite{taheri2020grab}.} 
We randomly randomly sample a manipulation trajectory for each object. 
If its manipulation is extremely simple, we additionally sample one trajectory for it. 
The object is not considered if its corresponding manipulation is bimanual such as \texttt{binoculars}, involves other body parts such as \texttt{bowl}, or with detailed part movements such as the \texttt{game} \texttt{controller}.
Finally, manipulations with the following objects are included in our dataset: 
\texttt{mouse}, \texttt{flashlight}, \texttt{stapler}, \texttt{hammer}, \texttt{torus}, \texttt{stanfordbunny}, \texttt{pyramid}, \texttt{cylinder}, \texttt{airplane}, \texttt{train}, \texttt{mouse} (resampled), \texttt{cube}, \texttt{watch}, \texttt{waterbottle}, \texttt{phone}, \texttt{sphere}, \texttt{mug}, \texttt{alarmclock}, \texttt{knife}, \texttt{fryingpan}, \texttt{cup}, \texttt{duck}, \texttt{elephant}, \texttt{lightbulb}, \texttt{scissors}, \texttt{toothbrush}, \texttt{toothpaste}. 
For each sequence, we take the first approach-action clip with the length of 60 frames. 
The number of manipulation sequences from GRAB is 27.

\begin{figure}[htbp]
  \centering
  \includegraphics[width=0.7\textwidth]{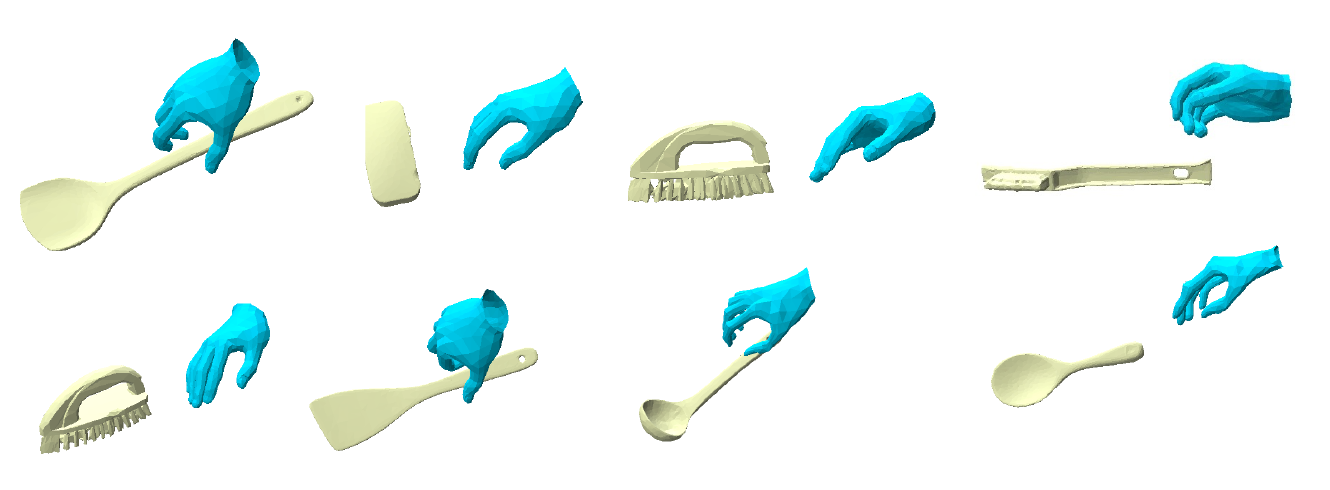}
  \caption{
  Snapshots from the TACO dataset.
  }
  \label{taco_data_sample}
\end{figure}

\noindent\textbf{TACO~\cite{liu2024taco}.} 
For TACO, we acquire data by contacting authors. 
We randomly select one sequence for each right-hand tool object, a few snapshots are presented in Fig.~\ref{taco_data_sample}.  Sequences with very low quality like erroneous object motions are excluded. For each trajectory, we take the first approach-action clip with the maximal length set to 150 frames. 
14 trajectories in total are selected finally.

\begin{table*}[t]
    \centering
    \caption{ 
    Default \textit{parameter settings} of the \textbf{quasi-physical simulator curriculum}.   
    } 
    \resizebox{1.0\textwidth}{!}{%
    \begin{tabular}{@{\;}lcccccccccc@{\;}}
        \toprule
        
        Object & \texttt{box} & \texttt{capsulemachine} & \texttt{espressomachine} & \texttt{ketchup}  &  \texttt{laptop}  & \texttt{microwave} & \texttt{mixer} & \texttt{phone} & \texttt{scissors} & \texttt{waffleiron}   \\

        \cmidrule(l{0pt}r{1pt}){1-1}
        \cmidrule(l{2pt}r{2pt}){2-11}

        Subject ID & 1 & 5 & 6 & 7 & 4 & 1 & 5 & 7 & 4 & 2
        \\

        \bottomrule
 
    \end{tabular}
    }
    \label{tb_exp_arctic_data}
\end{table*} 

\noindent\textbf{ARCTIC~\cite{fan2023arctic}.} 
For ARCTIC, we randomly select one sequence for each object from its available manipulation trajectories, resulting in 10 sequences in total. For each trajectory, we take the first approach-action clip with the maximal length set to 150 frames. The object names and the corresponding subject indexes are summarized in Table~\ref{tb_exp_arctic_data}. Please note that subject \texttt{s08} and \texttt{s09} only have ``\texttt{use}'' actions. Besides, some ``\texttt{grab}'' sequences are missing in a specific subject's manipulation sequences. For instance, both \texttt{s01} and \texttt{s06} do not have ``\texttt{grab}'' manipulations with \texttt{box}. 



\begin{figure}[htbp]
  \centering
  \includegraphics[width=0.7\textwidth]{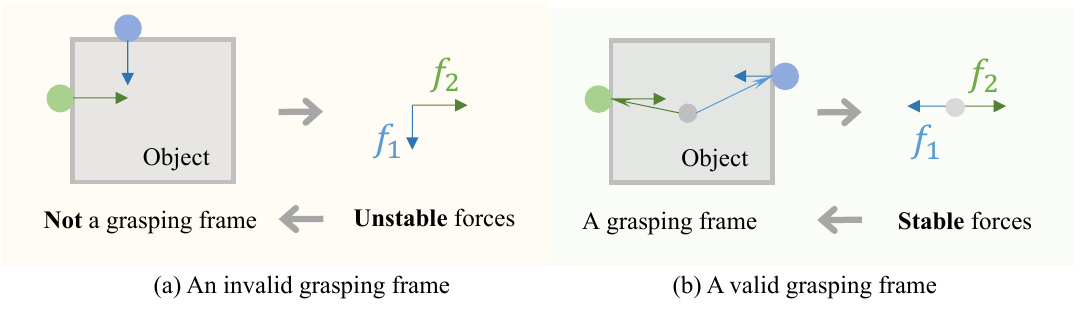}
  \caption{
  \textbf{Grasping frame.} We leverage a simple strategy to find the first grasping frame from the sequence. A valid grasping frame should have at least two contact points. The contact force directions should be able to stabilize the object, \emph{i.e.,} there exists a solution for their magnitudes so that zero force and zero torque are applied to the object. 
  }
  \label{fig_exp_grasping_frames}
\end{figure}

\subsection{Baseline}

\noindent\textbf{DGrasp-Base~\cite{christen2022d}.} 
We use the official code provided by authors\footnote{\href{https://github.com/christsa/dgrasp}{DGrasp's GitHub Repository.}}. 
We adapt the codebase to two simulated environments used in our evaluation, Bullet~\cite{coumans2016pybullet} and Isaac Gym~\cite{makoviychuk2021isaac}. 
Using DGrasp's method to complete the tracking task requires us to define reference grasping frames. 
We leverage a heuristic method and take the first grasp frame as the reference frame, illustrated in Fig.~\ref{fig_exp_grasping_frames}. Specifically, the first grasp is the first frame in the sequence satisfying the following conditions: 1) at least two contacts are detected between the hand and the object, 2) all contact force directions can form a force closure, that is there exists a solution for their magnitudes so that the object is stable under such contact forces. Having defined the reference grasping frame, we train the manipulation policy using the original DGrasp's method. 
Initially, only the grasping policy is activated. The grasping module guides the hand towards the object to find a stable grasp according to the defined reference frame. After that, the grasping policy and the control policy cooperate to move the object to the final 6D pose. Our method can find a successful policy on DGrasp's ``021\_bleach\_dexycb'' example in two simulated environments using the dynamic MANO hand~\cite{christen2022d}. 

\noindent\textbf{DGrasp-Tracking (improved from DGrasp~\cite{christen2022d}).} We set a series of reference frames from the sequence, where every two reference frames are separated by 10 frames. We use the grasping policy to guide the hand toward each reference frame.

\noindent\textbf{DGrasp-Tracking (w/ curric.).} We gradually train DGrasp-Tracking in each of the simulators from the quasi-physical simulators, finally in the tested simulator. The curriculum setting is the same as that listed in Table~\ref{tb_exp_curriculum_setting}.

\noindent\textbf{ControlVAE~\cite{yao2022controlvae}.} 
We adapt the official release\footnote{\href{https://github.com/heyuanYao-pku/Control-VAE}{ControlVAE's GitHub Repository.}} to the manipulation scenario. The world model approximates state transitions. It takes the current state, composed of the articulated dexterous robot hand joint state (including the first 6-DoF global rotations and translations), the object state, including the 4-dim object orientation represented as a quaternion, and the 3-DoF object translation, and control signals, including the velocity and position controls for each hand joint, as input. It outputs the predicted delta hand joint states and the predicted object delta rotations (3-DoF) as well as the delta translations (3-DoF). Following ControlVAE~\cite{yao2022controlvae}, the world model is an MLP. We increase the network depth, resulting in an MLP with 9 layers in total. The first hidden dimension is 256, followed by 6 layers with the hidden dimension of 512, 1 layer with the hidden dimension of 256, and the output layer. \texttt{ReLU} is used as the activation function between each hidden layer. 
The policy network takes the current state, including the hand joint state, object orientation as well as object rotation, and the target state, including the target hand joint states, target object orientation as well as the target object rotation as input. It predicts control signals for the articulated hand, including the position and velocity controls for each hand joint. 
The policy network is an MLP. The number of layers and the hidden dimension settings are the same as the world model. 
Length of the replay buffer is set to 1024. 
For Bullet, the batch size is set to 1. At each training loop, the world model is trained for 256 steps, followed by training the policy network for 256 steps. 
For Isaac Gym, the batch size is set to 128. At each training loop, the world model is trained for 8 steps, followed by training the policy network for 8 steps. 
Rollout lengths for the world model and for the policy are 24 and 19 respectively. 
The number of the maximum training iterations is set to 30000. 


\noindent\textbf{MPC (w/ base sim.).} 
The base simulator is the final analytical part of the quasi-physical simulator of the physics curriculum. Articulated rigid constraints are imposed. The spring-damper contact model is tuned to the stiffest level. 
Please refer to Table~\ref{tb_exp_curriculum_setting} for the setting of this simulator. 

\noindent\textbf{MPC (w/ base sim. w/ soften).}  
Based on the base simulator, we introduce the soften strategy present in Bundled Gradients~\cite{suh2022bundled}.
Penalty-based contacts are smoothed by sampling contact spring coefficients, as stated in Section IV.B~\cite{suh2022bundled}. The sampling range for each coefficient is defined as the [-10\%, +10\%] interval of the original value. 

\subsection{Experimental Settings}


\begin{table*}[t]
    \centering
    \caption{ 
    Default \textit{parameter settings} of the \textbf{quasi-physical simulator curriculum}.   
    } 
    \resizebox{1.0\textwidth}{!}{%
    \begin{tabular}{@{\;}lcccccccccc@{\;}}
        \toprule
        
        Simulator ID & 1 & 2 & 3 & 4  &  5  & 6 & 7 & 8 & 9 & 10   \\

        \cmidrule(l{0pt}r{1pt}){1-1}
        \cmidrule(l{2pt}r{2pt}){2-11}

        \makecell[l]{Point Set \\ Parameter $\alpha$} & 0.1 & 0 & 0 & 0 & 0 & 0 & 0 & 0 & 0 & 0
        \\ 
        \makecell[l]{Contact Distance \\ Threshold $d^c$} & 0.1 & 0.1 & 0.05 & 0.03 & 0.025 & 0.02 & 0.015 & 0.01 & 0.0 & 0.0
        \\ 
        \makecell[l]{Contact Spring \\ Stiffness $k^n$} & $4\times 10^{4}$ & $4\times 10^{4}$ & $8\times 10^{4}$ & $1\times 10^{5}$ & $2\times 10^{5}$ & $3\times 10^{5}$ & $3.5\times 10^{5}$ & $4\times 10^{5}$ & $4\times 10^{6}$ & $4\times 10^{6}$
        \\ 
        \makecell[l]{Friction Spring \\ Stiffness $k^f$} & $1\times 10^{5}$ & $1\times 10^{5}$ & $2\times 10^{5}$ & $4\times 10^{5}$ & $5\times 10^{5}$ & $6\times 10^{5}$ & $8\times 10^{5}$ & $1\times 10^{6}$ & $1\times 10^{7}$ & $1\times 10^{7}$
        \\ 
        \makecell[l]{Contact Damping \\ Coefficient $k^d$} & $1\times 10^{3}$ &$1\times 10^{3}$  & $1\times 10^{3}$  & $1\times 10^{3}$  & $1\times 10^{3}$  & $1\times 10^{3}$  & $1\times 10^{3}$  & $1\times 10^{3}$  & $1\times 10^{3}$  & $1\times 10^{3}$ 
        \\ 
        \makecell[l]{w/ Residual Physics?} & No & No & No & No & No & No & No & No & No & Yes
        \\ 

        \bottomrule
 
    \end{tabular}
    }
    \label{tb_exp_curriculum_setting}
\end{table*}

\begin{table*}[t]
    \centering
    \caption{ 
    \textit{Curriculum parameter settings} used in the ablated version (Ours w/ Curriculum II). 
    } 
    \resizebox{0.7\textwidth}{!}{%
    \begin{tabular}{@{\;}lccccccc@{\;}}
        \toprule
        
        Simulator ID & 1 & 2 & 3 & 4  &  5  & 6 & 7    \\

        \cmidrule(l{0pt}r{1pt}){1-1}
        \cmidrule(l{2pt}r{2pt}){2-8}

        \makecell[l]{Point Set \\ Parameter $\alpha$} & 0.1 & 0 & 0 & 0 & 0 & 0 & 0 
        \\ 
        \makecell[l]{Contact Distance \\ Threshold $d^c$} & 0.1 & 0.1 & 0.05 & 0.02 & 0.01 & 0.0 & 0.0
        \\ 
        \makecell[l]{Contact Spring \\ Stiffness $k^n$} & $4\times 10^{4}$ & $4\times 10^{4}$ & $8\times 10^{4}$ & $3\times 10^{5}$ & $4\times 10^{5}$ & $4\times 10^{6}$ & $4\times 10^{6}$
        \\ 
        \makecell[l]{Friction Spring \\ Stiffness $k^f$} & $1\times 10^{5}$ & $1\times 10^{5}$ & $2\times 10^{5}$ & $6\times 10^{5}$ & $1\times 10^{6}$ & $1\times 10^{7}$ & $1\times 10^{7}$
        \\ 
        \makecell[l]{Contact Damping \\ Coefficient $k^d$} & $1\times 10^{3}$ &$1\times 10^{3}$  & $1\times 10^{3}$  & $1\times 10^{3}$   & $1\times 10^{3}$  & $1\times 10^{3}$  & $1\times 10^{3}$ 
        \\ 
        \makecell[l]{w/ Residual Physics?} & No & No & No & No & No & No  & Yes
        \\ 

        \bottomrule
 
    \end{tabular}
    }
    \label{tb_exp_curriculum_setting_v2}
\end{table*}


\noindent\textbf{The quasi-physical simulator curriculum.} 
By default, the curriculum is composed of ten parameterized quasi-physical simulators. 
We summarize their parameter settings in Table~\ref{tb_exp_curriculum_setting}. 
The contact distance threshold $d^c$, contact spring stiffness $k^n$, friction spring stiffness $k^f$, and contact damping coefficient $k^d$ are set empirically. 

For the ablated version (``Ours w/ Curriculum II'' in the ablation study), we remove some stages from the original curriculum. The setting is summarized in Table~\ref{tb_exp_curriculum_setting_v2}. 

\noindent\textbf{Quasi-physical simulators.}
We use \texttt{Python} to implement each component of the simulator and the simulation processes, including the articulated rigid dynamics, the point set dynamics, the spring-damper contact modeling, and the residual physics modules. Semi-implicit time-stepping is leveraged. Time stepping is set to $5\times 10^{-4}$ with 100 substeps per frame.
In this way, we can easily introduce neural network components into the simulator. Besides, one can easily integrate it into 
a deep learning framework for further applications. Moreover, we can calculate gradients automatically taking advantage of the auto-grading feature of the framework. 
The overall efficiency, though has a large improvement space, is acceptable in our task. 

\noindent\textbf{Converting meshes to SDFs.} We use Mesh2SDF~\cite{mesh2sdf} in this process.

\noindent\textbf{Parameters set $\mathcal{S}$.} The parameter set $\mathcal{S}$ includes object properties, \emph{i.e.,} object mass and object inertia, and some unknown system parameters, \emph{i.e.,} linear velocity sampling coefficient and angular velocity damping coefficient. For the friction coefficient, we set it to a fixed value, \emph{i.e.,} $\mu = 10$. The value is set under the consideration of the important role friction forces play in the manipulation task. 

\noindent\textbf{Controlling the hand in Bullet and Isaac Gym.} 
In our quasi-physical simulator, the hand is controlled via joint forces and root linear and angular velocities. In Bullet and Isaac Gym, people commonly use PD controls, which are also recommended officially~\cite{coumans2016pybullet}. Therefore, to convert controls in joint forces and root velocities to PD controls in the them, we additionally add a control transformation module. 

For each timestep $1\le n\le N-1$, it takes root positions at the timestep $n$ and $n+1$, joint states, velocities, joint forces, and the object state at step $n$ and outputs the residual position and velocity controls at step $n$. The predicted residual PD controls added to the root positions, root velocities (calculated via finite differences), joint states, and velocities are treated as PD controls in the target simulator. 
The control transformation module is composed of a hand point feature extraction layer, an object feature extraction layer, and a prediction layer. 
The current hand and object geometry is firstly encoded in latent features. Subsequently, the original joint control related information and the encoded latent features are fed into an MLP for residual position and velocity control prediction. 
The feature extraction layer is a 3-layer MLP with hidden dimensions [128, 128, 128] and \texttt{ReLU} as the activation layer. After per-point feature extraction, a \texttt{maxpool} function operates on point features to extract global features for the hand and the object. Then the global features of the hand and the object are concatenated together and passed through a two-layer MLP with hidden dimension 128 and the output dimension 128 as well. The output feature is then concatenated with the object control related information and passed through an MLP for the residual control prediction. The prediction network is a 3-layer MLP with hidden dimension [128, 64]. 
The control transformation module is optimized together with the residual physics module introduced in the parameterized quasi-physical simulator. 

\noindent\textbf{World model-style training. }
Rollout lengths for both the trajectory optimization and the model training are set to 19. In each iterative training iteration, the trajectory is optimized for 256 steps. The residual physics module is optimized for 256 steps. The replay buffer length is 1024.

\noindent\textbf{Evaluation process.} 
Our method is a multi-stage optimization-based strategy. 
The overall optimization process can be roughly divided into three stages, as illustrated in the following:
\begin{itemize}
    \item \textbf{Transferring via point dynamics.} This stage involves three processes: 
    \begin{itemize}
        \item Optimize a dynamics MANO~\cite{christen2022d} trajectory that can track the input kinematics-only trajectory;
        \item Optimize the control trajectory for the point set of the MANO hand that can track the hand trajectory and  the object trajectory;
        \item Optimize a kinematics-only trajectory for the simulated robot hand so that it can track the kinematic hand trajectory via sparse correspondences;
        \item Optimize the control trajectory for the point set of the simulated robot hand so that it can track the trajectory of the MANO's point set. 
    \end{itemize}
    \item \textbf{Optimizing through a contact curriculum.} In this stage, the control trajectory of the simulated robot hand is optimized in each simulator from the curriculum. The objective is to track the hand trajectory and the object trajectory.
    \item  \textbf{Transferring to a realistic simulated environment.} In this stage, the quasi-physical simulator and the control trajectory for the simulated robot hand are iteratively optimized. By default, the number of iterations is set to 30,000. 
\end{itemize}
In each optimization iteration, excluding the kinematics trajectory-only optimization, the parameter set $\mathcal{S}$ and the control trajectory are optimized alternately. If we cannot inherit a control trajectory from the previous stage, we first optimize the it with the parameters $\mathcal{S}$ either inherited from previous stages or set to default values. After that, the parameters $\mathcal{s}$ are further refined with controls fixed. Subsequently, we continue to optimize controls based on the identified parameters. If the control trajectory can be inherited from previous stages at the beginning of the iteration, the parameters $\mathcal{S}$ are identified with controls fixed. Then we further refine controls with parameters fixed. Typically, the number of optimization steps for the parameters is 1000, while the number is 100 for the control trajectory. 
Both hand controls and parameters are optimized via gradient descent. 
Learning rate is set to $5\times 10^{-4}$ for both control optimization and parameters identification. 
We use \texttt{Adam} optimizer. No learning rate scheduler is used. 
In the third stage, we follow the training framework in ControlVAE~\cite{yao2022controlvae}. The optimizer is \texttt{RAdam}, with the learning rate $10^{-4}$ for the quasi-physical simulator and $10^{-4}$ for control trajectory optimization. 



%
\subsection{Running Time and Complexity} 

\noindent\textbf{Complexity.} 
The time complexity is related to the number of frames in the manipulation sequence and the number of optimization passes. Denote the number of frames as $N$ and the number of total optimization passes as $K$, the time complexity is $\mathcal{O}(KN)$. 

\noindent\textbf{Running time.} 
Taking a sequence with 60 frames as an example, the first stage (see \textbf{evaluation process} stated in the previous section) costs about 7 hours in total. 
Using the default curriculum setting (Table~\ref{tb_exp_curriculum_setting}), the second stage would cost about 22 hours. Early termination logic in each optimization iteration will shorten the time. Therefore, the actual time is per-sequence dependent. Taking transferring to the Bullet simulator as an example, the third stage takes about 20 hours to complete. 
Reducing the number of simulators in the curriculum or using a smaller number of iterations in the third stage can improve the time efficiency.


\section{Potential Negative Societal Impact} \label{sec:neg_social_impact}
Our approach has the potential to expedite the advancement of robotic dexterous manipulation skills. However, in the future, the emergence of highly developed robots proficient in performing various tasks may lead to the replacement of certain human labor, thus potentially impacting society.

\end{document}